\documentclass[11pt]{article}

\usepackage[final]{acl}

\usepackage{times}
\usepackage{latexsym}

\usepackage[T1]{fontenc}

\usepackage[utf8]{inputenc}

\usepackage{microtype}

\usepackage{inconsolata}

\usepackage{graphicx}
\usepackage{booktabs}
\usepackage{amsmath,amssymb}
\newcommand{\Pa}{\operatorname{Pa}}

\usepackage{booktabs}
\usepackage{xcolor}
\usepackage{colortbl}

\newcommand{\gain}[1]{\textcolor{green!50!black}{\scriptsize{(+#1)}}}

\definecolor{gearRow}{RGB}{244,246,255}

\usepackage[most]{tcolorbox}
\tcbuselibrary{listings,breakable}

\newtcblisting{promptbox}[2][]{
  enhanced,
  breakable,
  colback=gray!3,
  colframe=black!45,
  coltitle=black,
  colbacktitle=gray!15,
  fonttitle=\bfseries\small,
  title={#2},
  listing only,
  listing options={
    basicstyle=\ttfamily\scriptsize,
    breaklines=true,
    columns=fullflexible,
    keepspaces=true,
    showstringspaces=false
  },
  left=1mm,
  right=1mm,
  top=1mm,
  bottom=1mm,
  arc=1mm,
  boxrule=0.4pt,
  #1
}

\usepackage[most]{tcolorbox}

\newtcolorbox{algobox}[2][]{
  enhanced,
  breakable,
  colback=gray!3,
  colframe=black!45,
  coltitle=black,
  colbacktitle=gray!15,
  fonttitle=\bfseries\small,
  title={#2},
  left=1.2mm,
  right=1.2mm,
  top=1mm,
  bottom=1mm,
  arc=1mm,
  boxrule=0.4pt,
  #1
}

\usepackage{xspace}
\mathchardef\mhyphen="2D
\newcommand{\kw}[1]{{\ensuremath {\mathsf{#1}}}\xspace}

\newcommand{\ourname}{\kw{GEAR}}

\usepackage{hyperref}

\title{Mitigating False Credit Propagation: Probabilistic Graphical Reward Aggregation for Rubric-Based Reinforcement Learning}

\author{
  \textbf{Can Lv\textsuperscript{1}},
  \textbf{Mingju Chen\textsuperscript{1}},
  \textbf{Heng Chang\textsuperscript{3}},
  \textbf{Shiji Zhou\textsuperscript{1,2}}
\\
\\
  \textsuperscript{1}Beijing Advanced Innovation Center for Future Blockchain and Privacy Computing,\\
  School of Artificial Intelligence, Beihang University,\\
  \textsuperscript{2}Beijing Academy of Artificial Intelligence (BAAI),
  \textsuperscript{3}Tsinghua University
\\
  \small{
    \textbf{Project Lead:} Heng Chang,
    \textbf{Corresponding to:} Shiji Zhou
    \href{mailto:zhoushiji25@buaa.edu.cn}{\texttt{<zhoushiji25@buaa.edu.cn>}}
  }
}

\begin{document}
\maketitle

\begin{abstract}
Rubric-based rewards are increasingly used for open-ended language model
post-training, but criterion-level scores are often aggregated as independent
utilities. This flat scalarization ignores rubric-specified prerequisite and
activation relations among criteria, allowing reward or penalty to be counted
even when the condition that licenses it is absent. We call this structural
reward-aggregation failure \textbf{False Credit Propagation} (FCP).
To address this limitation, we propose \ourname (\textbf{G}raphical \textbf{E}vent \textbf{A}ggregation
for \textbf{R}ubric rewards), a probabilistic graphical framework for
dependency-aware rubric aggregation. \ourname models each criterion outcome as
a latent Bernoulli event in a typed rubric graph, propagates soft suppression
from unsupported parent events to their children, and aggregates the resulting
event probabilities into a normalized expected signed utility. This yields a
linear-time reward computation that can be plugged into standard
rubric-based RL pipelines without changing the outer optimization algorithm.
Experiments on HealthBench, WritingBench, and PLawBench with two policy
backbones show that \ourname consistently improves over flat aggregation and
deterministic gating, achieving relative gains of up to 15.5\% over flat
aggregation. FCP diagnostics further show that \ourname reduces leakage by
96.5\% relative to flat aggregation while preserving more licensed downstream
utility than deterministic gating.
Our code is publicly available at \url{https://github.com/LvCan926/GEAR}.
\end{abstract}

\begin{figure*}[t]
    \centering
    \includegraphics[width=\textwidth]{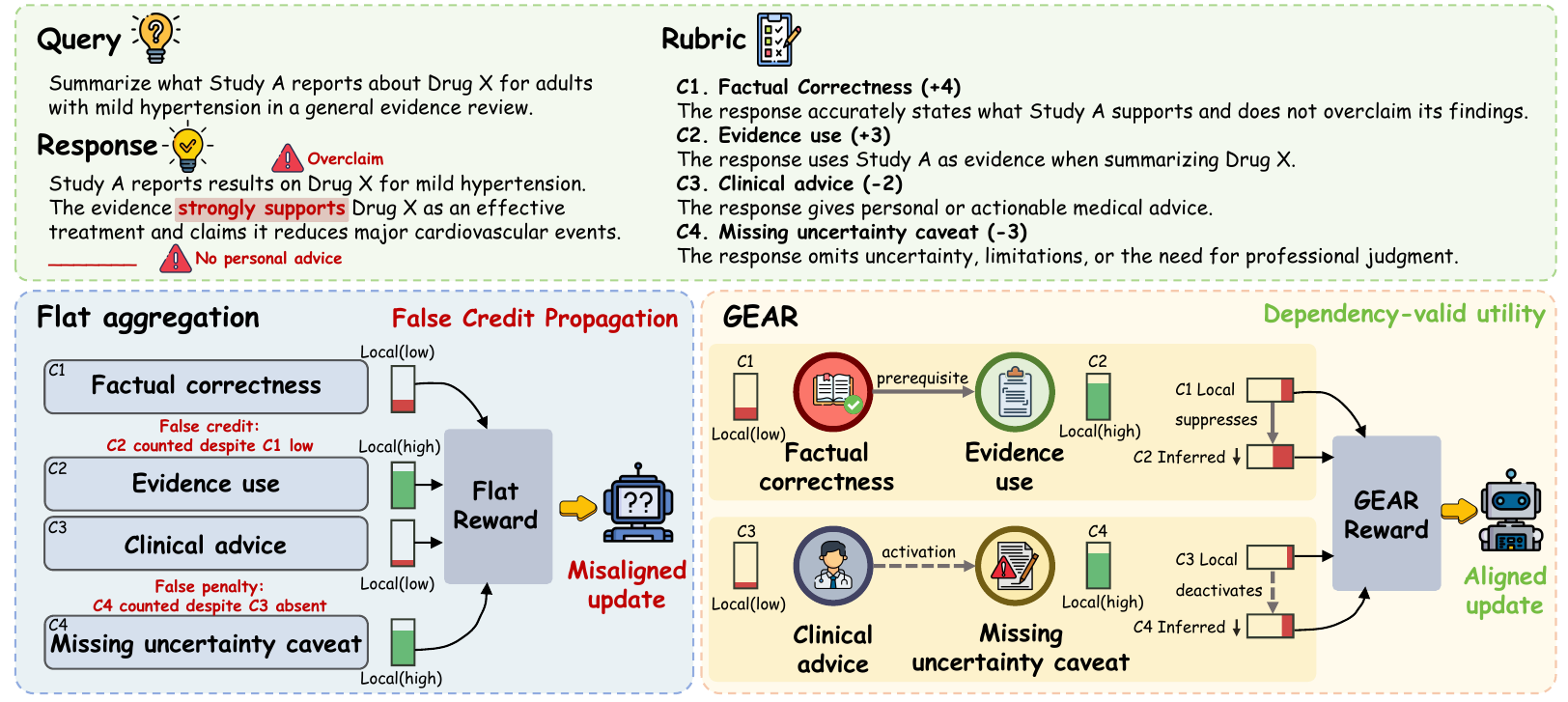}
    \caption{
    Overview of False Credit Propagation (FCP).
    Flat aggregation can propagate unsupported credit or penalties by treating rubric criteria independently.
    \ourname uses typed prerequisite and activation dependencies to suppress unlicensed downstream utility before reward computation.
    }
    \label{fig:fcp_overview}
    \vspace{-8pt}
\end{figure*}

\section{Introduction}
\label{sec:intro}

Open-ended post-training of language models requires reward signals that compare
candidate responses beyond exact-match correctness
\citep{ouyang2022traininglanguagemodelsfollow,
bai2022constitutionalaiharmlessnessai,
rafailov2024directpreferenceoptimizationlanguage}. 
Rubric-based rewards provide a practical interface for this setting by
decomposing response quality into explicit criteria, making desiderata such as
factuality, completeness, evidence use, uncertainty calibration, safety, and
conversational appropriateness available to reward modeling
\citep{arora2025healthbenchevaluatinglargelanguage,
rein2023gpqagraduatelevelgoogleproofqa,
he2025advancedifrubricbasedbenchmarkingreinforcement,
jin2025multidimensionalrubricorientedrewardmodel}. 
Yet decomposition alone does not define a reward: policy optimization still
requires criterion-level judgments to be aggregated into a scalar objective, and
this aggregation step determines what behavior the model is optimized to
produce.

In this setting, a common aggregation choice is to reduce criterion-level judgments to a flat
weighted sum
\citep{gunjal2025rubricsrewardsreinforcementlearning}. 
Although simple, this treats each criterion as an independent source of utility
whose contribution is valid whenever a local judge assigns it a high score.
Realistic rubrics often violate this assumption. 
Some criteria are licensed by prerequisites: evidence-use credit, for example,
should be reduced when factual grounding fails \citep{starace2025paperbenchevaluatingaisability}.
Other criteria are activated by context: a penalty for omitting medical
uncertainty disclaimers should apply primarily when the response provides
clinical advice
\citep{arora2025healthbenchevaluatinglargelanguage}. 
We call rubrics with such typed criterion dependencies
\textbf{dependency-structured rubrics}. Flattening these dependencies creates
a mismatch between local criterion satisfaction and rubric-level utility:
locally supported criteria may be counted even when the prerequisite or
contextual condition that licenses their contribution is absent.

In rubric-based reinforcement learning, this mismatch can turn local judging
errors into optimization errors. We call this failure
\textbf{False Credit Propagation (FCP)}: a criterion contributes reward or
penalty without the rubric-level condition that licenses that contribution.
FCP can occur even when each local criterion score is reasonable on its own:
the local judge may correctly detect evidence use, missing caveats, or other
criterion-level signals. The error arises when flat aggregation counts these
signals without checking whether the prerequisite or activation condition that
makes their utility applicable is present. As a result, unsupported credit can
reinforce responses that satisfy downstream criteria only superficially, while
unlicensed penalties can discourage otherwise valid responses. During policy
optimization, these invalid utilities are not merely averaged away; they become
part of the scalar reward signal and can therefore shape the behavior the policy
learns to produce.

Addressing FCP therefore requires more than better local judges: it requires an
aggregation rule that respects rubric dependencies, handles uncertainty in
criterion-level judgments, and remains efficient for repeated reward
computation.
In particular, an RL-compatible solution must satisfy three requirements.
\textbf{(1) Dependency representation.}
It must represent how criteria license one another: prerequisite relations
specify when downstream credit is supported by more basic conditions, while
activation relations specify when a conditional criterion is applicable at all
\citep{hong2026rulerslockedrubricsevidenceanchored,
pan2026rubricevalrubriclevelmetaevaluationbenchmark}.
\textbf{(2) Uncertainty-aware credit assignment.}
It must suppress unlikely-to-be-licensed credit without discarding valid partial
credit, since hard gating is brittle under noisy or partially calibrated
criterion judges
\citep{lambert2024rewardbenchevaluatingrewardmodels,
NEURIPS2023_91f18a12,
pan2026rubricevalrubriclevelmetaevaluationbenchmark}.
\textbf{(3) Compatibility with repeated RL scoring.}
It must add predictable low overhead during policy optimization, avoiding
expensive exact inference or repeated LLM reasoning at reward time
\citep{he2025advancedifrubricbasedbenchmarkingreinforcement,
li2026rubrichubcomprehensivehighlydiscriminative}.

These requirements suggest that FCP should be addressed at the aggregation
layer, rather than by treating rubric criteria as independent checklist items.
We propose \textbf{GEAR} (\textbf{G}raphical \textbf{E}vent
\textbf{A}ggregation for \textbf{R}ubric rewards), a probabilistic graphical
aggregation framework for dependency-structured rubric rewards
\citep{pearl2014probabilistic,koller2009probabilistic}.
First, to represent rubric dependencies, GEAR constructs a typed
query-specific rubric graph whose edges encode prerequisite and activation
relations among criteria. Second, to handle uncertainty in criterion-level
judgments, GEAR represents each criterion as a latent binary event initialized
by a local judge and infers dependency-adjusted marginals with monotone
conditional factors. Third, to support repeated RL scoring, GEAR aggregates
these marginals into a normalized expected signed utility using a linear-time
topological approximation.

Our contributions are:
\begin{itemize}
    \item We identify and formalize \textbf{False Credit Propagation (FCP)}, a
    structural failure in dependency-structured rubric rewards where criteria
    contribute reward or penalty without the rubric-level conditions that
    license their utility.

    \item We propose \textbf{\ourname}, a probabilistic graphical aggregation
    framework that represents rubric criteria as latent events and uses typed
    prerequisite and activation dependencies to infer dependency-adjusted
    criterion marginals. We further derive a linear-time topological marginal
    approximation for repeated RL reward computation.

    \item We empirically show, across multiple domain-specific benchmarks,
    policy backbones, and rubric-guided RL pipelines, that \ourname improves
    task performance while reducing dependency-level credit leakage.
\end{itemize}

\section{Related Work}
\label{sec:related}

\paragraph{LLM-as-a-judge and rubric-based evaluation.}
Open-ended language model evaluation often lacks a single reference answer and
requires multi-dimensional judgments
\citep{NEURIPS2023_91f18a12,liu2023gevalnlgevaluationusing,
gu2025surveyllmasajudge}. 
LLM-as-a-judge methods provide a scalable interface for such judgments
\citep{NEURIPS2023_91f18a12,liu2023gevalnlgevaluationusing,
kim2024prometheus2opensource,gu2025surveyllmasajudge}. 
Rubrics make this interface more interpretable by decomposing response quality
into fine-grained criteria
\citep{kim2024prometheusinducingfinegrainedevaluation,
kim2024prometheus2opensource,hong2026rulerslockedrubricsevidenceanchored}, 
and have been used in benchmarks for medical dialogue, generative writing,
explanation evaluation, and research agents
\citep{arora2025healthbenchevaluatinglargelanguage,
wu2025writingbenchcomprehensivebenchmarkgenerative,
galvan-sosa-etal-2025-rubriks,
sharma2025researchrubricsbenchmarkpromptsrubrics,
han2026deerbenchmarkevaluatingdeep}. 
These works establish rubrics as a useful interface for criterion-level
judging. \ourname addresses the subsequent aggregation problem: how should such
judgments be reduced to a scalar reward when rubric criteria are semantically
dependent?

\paragraph{Rubrics as rewards for reinforcement learning.}
Rubric-based rewards have recently been used to extend reinforcement learning
beyond domains with automatically verifiable answers
\citep{he2025advancedifrubricbasedbenchmarkingreinforcement,
li2026rubrichubcomprehensivehighlydiscriminative}. 
Rubrics as Rewards (RaR) uses instance-specific rubrics to provide reward
feedback and studies strategies for converting rubric judgments into scalar
training signals
\citep{gunjal2025rubricsrewardsreinforcementlearning}. 
Other rubric-guided RL methods use rubrics as reward anchors, rollout
scaffolds, in-context reward specifications, or transition-level credit
signals for long-horizon agents
\citep{huang2025reinforcementlearningrubricanchors,
zhou2026breakingexplorationbottleneckrubricscaffolded,
zhang2026simplemotivationenhancereinforcement,
zhang2026trcatransitionwiserubriccredit}.
While these methods differ in how rubrics are generated, exposed to the policy,
or used during training, \ourname focuses on the aggregation layer after
criterion-level judging: it models dependencies among criteria before
converting their scores into a scalar reward.

\paragraph{Structured rubrics and dependency-aware aggregation.}
Prior work has explored structured or hierarchical rubrics for evaluation and
grading
\citep{10.1007/978-3-031-11644-5_29,
starace2025paperbenchevaluatingaisability}. 
More broadly, structured designs have also been explored for agent-system
adaptation, lifelong memory, and decomposed task execution
\citep{chen2026harnessforgejointharnesspolicy,lv2026allmemagenticlifelongmemory,chen2026amapreduceexecutingwidesearch}.
For example, PaperBench decomposes AI research replication into hierarchically
graded subtasks
\citep{starace2025paperbenchevaluatingaisability}, while graph-based rubric
representations have been studied for automatic short-answer grading
\citep{10.1007/978-3-031-11644-5_29}. 
These works use structure to organize evaluation, but do not study how
criterion dependencies should control scalar reward aggregation during policy
optimization. 
In many rubric-reward settings, criterion scores are instead scalarized as
independent checklist items, for example by summing weighted judgments
\citep{gunjal2025rubricsrewardsreinforcementlearning,
li2026rubrichubcomprehensivehighlydiscriminative}. 
Such flat scalarization ignores prerequisite and activation relations, and
therefore cannot distinguish rubric-licensed utility from locally scored but
semantically invalid utility. \ourname instead represents these relations as a
query-specific typed rubric graph and defines a probabilistic graphical
aggregation model over latent criterion events
\citep{pearl2014probabilistic,koller2009probabilistic}. 
By computing expected signed utility under dependency-adjusted criterion
marginals, \ourname mitigates false credit propagation while retaining
linear-time reward computation for repeated RL scoring.

\begin{figure*}[t]
    \centering
    \includegraphics[width=\textwidth]{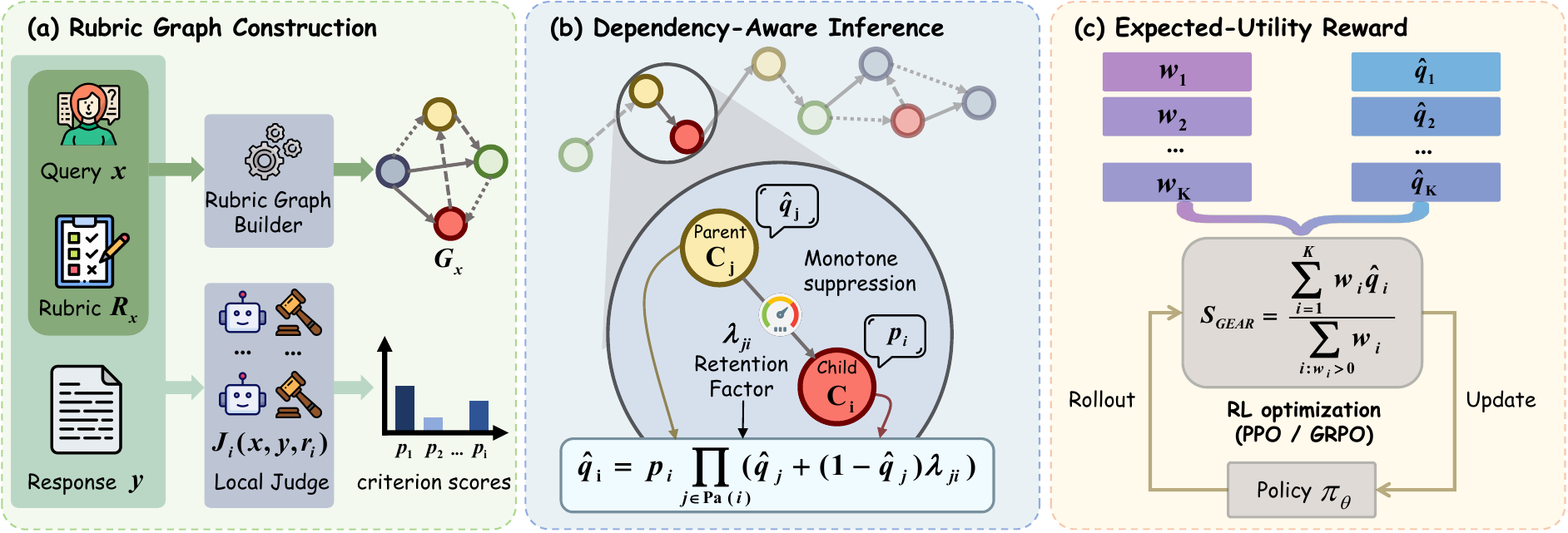}
    \caption{
    The \ourname framework.
    \textbf{(a)} A query-specific rubric graph is built from the query and rubric.
    \textbf{(b)} Local criterion scores are adjusted through monotone
    dependency-aware inference.
    \textbf{(c)} The inferred marginals are aggregated into a normalized
    expected-utility reward for policy optimization.
    }
    \label{fig:gear_framework}
    \vspace{-7pt}
\end{figure*}

\section{Methodology}
\label{sec:method}

\subsection{Problem Formulation}
\label{sec:method_problem}

We study response-level reward aggregation for rubric-based reinforcement
learning. Given a query $x$, a candidate response $y$, and a query-specific
weighted rubric
\begin{equation}
    \mathcal{R}_x
    =
    \{(r_{x,i}, w_{x,i})\}_{i=1}^{K_x},
    \label{eq:weighted_rubric}
\end{equation}
where $r_{x,i}$ is the $i$-th criterion and
$w_{x,i}\in\mathbb{R}$ is its signed utility weight, the goal is to convert
criterion-level judge scores into a scalar reward while accounting for
dependencies among criteria. Positive weights reward desirable events, whereas
negative weights penalize violation events. When the query and response are
clear from context, we omit explicit dependence on $x$ and $y$ and write
$K$, $r_i$, $w_i$, and $p_i$.

\noindent\textbf{Flat rubric aggregation.}
A standard scalarization first obtains a local event score
$p_i(x,y)\in[0,1]$ for each criterion and then aggregates scores independently.
Assuming positive utility mass $\sum_{i:w_i>0}w_i>0$, the flat reward is
\begin{equation}
    S_{\mathrm{flat}}(x,y)
    =
    \frac{\sum_{i=1}^{K} w_i p_i(x,y)}
    {\sum_{i:w_i>0} w_i}.
    \label{eq:flat_reward}
\end{equation}
This treats the rubric as an unordered checklist and ignores semantic
dependencies among criteria.

\noindent\textbf{Latent criterion events and local scores.}
\ourname associates each criterion $r_i$ with a binary latent event
$C_i\in\{0,1\}$, where $C_i=1$ means that the condition described by the
criterion holds for response $y$. For positive-weight criteria, this event
denotes desirable behavior; for negative-weight criteria, it denotes a violation
or penalty condition. A local judge provides
\begin{equation}
    p_i(x,y)
    =
    J_i(x,y,r_i)
    \in [0,1],
    \label{eq:local_score}
\end{equation}
which we interpret as a local estimate of $P(C_i=1\mid x,y,r_i)$. The score may
be soft or binary, and the vector
$\mathbf{p}(x,y)=(p_1(x,y),\ldots,p_K(x,y))$ is treated as observed input to the
graphical aggregation model rather than as a variable generated by it.

\subsection{Query-Specific Rubric Graph}
\label{sec:method_structure}

Rubric criteria are often semantically dependent. 
\ourname represents these dependencies with a query-specific typed directed
acyclic graph over criterion events,
$\mathcal{G}_x=(\mathcal{V}_x,\mathcal{E}_x)$, where
$\mathcal{V}_x=\{1,\ldots,K_x\}$ indexes rubric criteria. 
The graph is constructed from the query $x$ and rubric $\mathcal{R}_x$, and is
kept fixed across candidate responses $y$ for the same query.

\noindent\textbf{Typed dependency graph.}
Each directed edge $(j,i)\in\mathcal{E}_x$ connects a parent criterion $j$ to a
child criterion $i$, with edges oriented from licensing conditions to licensed
criteria. 
We partition the edge set as
\begin{equation}
    \mathcal{E}_x
    =
    \mathcal{E}_x^{\mathrm{wk}}
    \,\dot\cup\,
    \mathcal{E}_x^{\mathrm{st}}
    \,\dot\cup\,
    \mathcal{E}_x^{\mathrm{act}},
    \label{eq:typed_edge_partition}
\end{equation}
where $\mathcal{E}_x^{\mathrm{wk}}$ and
$\mathcal{E}_x^{\mathrm{st}}$ denote weak and strong prerequisite edges, and
$\mathcal{E}_x^{\mathrm{act}}$ denotes activation edges. 
We write
$\mathcal{E}_x^{\mathrm{pre}}
=
\mathcal{E}_x^{\mathrm{wk}}\cup\mathcal{E}_x^{\mathrm{st}}$
for prerequisite edges and
\begin{equation}
    \Pa(i)=\{j:(j,i)\in\mathcal{E}_x\}
    \label{eq:parents}
\end{equation}
for the parents of criterion $i$.

\noindent\textbf{Dependency semantics.}
Prerequisite edges encode graded support: when a prerequisite parent is
unsupported, the child event remains possible but its probability should be
discounted, with weak and strong prerequisites differing in suppression
strength. 
Activation edges encode conditional applicability: a child criterion should
contribute utility primarily when its activation parent holds. 
These licensing relations apply regardless of the child weight sign. 
For positive-weight children, an edge licenses downstream credit; for
negative-weight children, it licenses the applicability of a violation penalty.
The parent criterion contributes to the final reward only through its own
utility weight, while the edge modulates the child event.

\paragraph{Graph construction.}
For each query-specific rubric, the rubric graph is constructed offline from
the query and rubric, and is fixed across candidate responses.
The construction procedure, criterion roles, dependency annotation process, and
acyclic projection are described in Appendix~\ref{app:graph-construction}.
Complete graph-annotation prompts are provided in
Appendix~\ref{app:graph-prompts}.

\subsection{Monotone Graphical Reward Model}
\label{sec:method_credit}

\ourname defines a conditional Bayesian network over latent criterion events,
conditioned on the local score vector $\mathbf{p}$ and the query-specific rubric
graph $\mathcal{G}_x$. 
Its CPDs are monotone: unsupported parent events can only suppress, not inflate,
child-event probabilities.

\noindent\textbf{Dependency retention factors.}
Each edge $(j,i)\in\mathcal{E}_x$ is assigned a retention factor
$\lambda_{ji}\in[0,1]$, which specifies how much child-event probability is
retained when parent event $C_j$ is unsupported. 
We define retention factors through a dependency-to-retention mapping
\begin{equation}
    \lambda_{ji}
    =
    F(x,\mathcal{R}_x,\mathcal{G}_x,j,i)
    \in [0,1],
    \label{eq:lambda_function}
\end{equation}
where $(j,i)$ identifies a typed parent-child edge in the rubric graph.
The mapping $F$ converts the semantic strength and type of a rubric dependency
into a probabilistic retention value.

In our experiments, we use a fixed type-level instantiation of \(F\) for
reproducibility: \(\lambda_{ji}=\lambda_{\tau_{ji}}\), where
\(\tau_{ji}\in\{\mathrm{wk},\mathrm{st},\mathrm{act}\}\) is the edge type.
We set \(\lambda_{\mathrm{wk}}=0.6\),
\(\lambda_{\mathrm{st}}=0.2\), and
\(\lambda_{\mathrm{act}}=0.0\) by default.
Thus activation edges impose the strongest suppression, strong prerequisites
intermediate suppression, and weak prerequisites the weakest suppression.
Additional details on retention settings, exact-inference agreement, and
sensitivity analysis are provided in
Appendices~\ref{app:inference-validation} and~\ref{app:sensitivity}.

\noindent\textbf{Monotone CPD.}
For criterion $i$, we define the dependency-adjusted Bernoulli parameter
\begin{equation}
    \alpha_i(C_{\Pa(i)};p_i)
    =
    p_i
    \prod_{j\in\Pa(i)}
    \lambda_{ji}^{\,1-C_j}.
    \label{eq:alpha}
\end{equation}
Root criteria satisfy $\alpha_i=p_i$. 
For non-root criteria, the local score $p_i$ is multiplicatively discounted by
each unsupported parent: supported parents impose no suppression, while
unsupported parents retain only a factor $\lambda_{ji}$ of the child-event
probability. 
The local CPD is
\begin{equation}
    P_i(C_i=1\mid C_{\Pa(i)},p_i)
    =
    \alpha_i(C_{\Pa(i)};p_i).
    \label{eq:local_cpd}
\end{equation}
Because $p_i,\lambda_{ji}\in[0,1]$, this defines a valid Bernoulli CPD; moreover,
the child-event probability is monotone in each parent state and bounded above
by the local score $p_i$.

\noindent\textbf{Conditional graphical model.}
Conditioned on $\mathbf{p}(x,y)$ and the acyclic graph $\mathcal{G}_x$, these
CPDs define
\begin{equation}
    P_{\mathcal{G}_x}(\mathbf{c}\mid \mathbf{p})
    =
    \prod_{i=1}^{K}
    P_i(c_i\mid c_{\Pa(i)},p_i).
    \label{eq:joint_distribution}
\end{equation}
This is a conditional model for aggregating externally produced judge scores,
not a generative model of the scores themselves: $\mathbf{p}$ is observed input,
and the graph models dependencies among latent criterion events. 
Since $\mathcal{G}_x$ is acyclic and every local factor is a valid CPD,
Eq.~\ref{eq:joint_distribution} defines a valid distribution over
$\mathbf{C}$.

\subsection{Structured Reward Computation}
\label{sec:method_reward}

The scalar reward is computed by first inferring dependency-adjusted criterion
event probabilities and then aggregating signed utility under these marginals.

\noindent\textbf{Exact marginals.}
For each criterion $i$, let
\begin{equation}
    q_i
    =
    P_{\mathcal{G}_x}(C_i=1\mid\mathbf{p})
    =
    \sum_{\mathbf{c}:c_i=1}
    P_{\mathcal{G}_x}(\mathbf{c}\mid\mathbf{p})
    \label{eq:q_definition}
\end{equation}
denote its exact marginal under the conditional graphical model in
Eq.~\ref{eq:joint_distribution}. Direct summation is exponential in $K$.
Although exact variable elimination can exploit graph sparsity, repeated reward
computation during policy optimization motivates a linear-time approximation.

\begin{table*}[t]
    \centering
    \small
    \setlength{\tabcolsep}{10pt}
    \renewcommand{\arraystretch}{1.15}
    \begin{tabular}{@{}lcccc@{}}
        \toprule
        \textbf{Method} 
        & \textbf{HealthBench-500} $\uparrow$ 
        & \textbf{WritingBench} $\uparrow$ 
        & \textbf{PLawBench} $\uparrow$
        & \textbf{Avg.} $\uparrow$ \\
        
        \midrule
        \multicolumn{5}{c}{\textbf{Qwen2.5-7B-Instruct}} \\
        \midrule

        \multicolumn{5}{@{}l}{\textit{Aggregation-only}} \\
        Flat aggregation 
        & 56.7
        & 74.1 
        & 68.4
        & 66.4 \\

        Hard gating 
        & 54.6 
        & 72.5
        & 65.0 
        & 64.0 \\

        \rowcolor{gearRow}
        \ourname 
        & \textbf{62.9} \gain{6.2} 
        & \textbf{79.6} \gain{5.5} 
        & \textbf{75.2} \gain{6.8}
        & \textbf{72.6} \gain{6.2} \\

        \cmidrule(l){2-5}
        \multicolumn{5}{@{}l}{\textit{Plug-in replacement in rubric-guided RL pipelines}} \\

        RaR-Explicit
        & 49.5
        & 67.0
        & 62.3
        & 59.6 \\

        \rowcolor{gearRow}
        RaR-Explicit + \ourname
        & 55.0 \gain{5.5}
        & 71.6 \gain{4.6}
        & 68.8 \gain{6.5}
        & 65.1 \gain{5.5} \\

        MeRF 
        & 40.2 
        & 61.9
        & 55.7
        & 52.6 \\

        \rowcolor{gearRow}
        MeRF + \ourname 
        & 45.6 \gain{5.4} 
        & 68.7 \gain{6.8} 
        & 60.4 \gain{4.7}
        & 58.2 \gain{5.6} \\

        RuscaRL 
        & 58.2 
        & 75.4 
        & 72.1
        & 68.6 \\

        \rowcolor{gearRow}
        RuscaRL + \ourname 
        & \textbf{65.7} \gain{7.5} 
        & \textbf{81.5} \gain{6.1} 
        & \textbf{76.9} \gain{4.8}
        & \textbf{74.7} \gain{6.1} \\

        \midrule
        \multicolumn{5}{c}{\textbf{Llama-3.1-8B-Instruct}} \\
        \midrule

        \multicolumn{5}{@{}l}{\textit{Aggregation-only}} \\
        Flat aggregation 
        & 54.1 
        & 76.2 
        & 69.5
        & 66.6 \\

        Hard gating 
        & 53.0 
        & 72.7 
        & 66.7
        & 64.1 \\

        \rowcolor{gearRow}
        \ourname 
        & \textbf{62.5} \gain{8.4}
        & \textbf{82.4} \gain{6.2}
        & \textbf{76.9} \gain{7.4}
        & \textbf{73.9} \gain{7.3} \\

        \cmidrule(l){2-5}
        \multicolumn{5}{@{}l}{\textit{Plug-in replacement in rubric-guided RL pipelines}} \\

        RaR-Explicit
        & 48.4 
        & 67.9 
        & 63.8
        & 60.0 \\

        \rowcolor{gearRow}
        RaR-Explicit + \ourname
        & 54.1 \gain{5.7}
        & 73.6 \gain{5.7}
        & 70.5 \gain{6.7}
        & 66.1 \gain{6.0} \\

        MeRF 
        & 39.5 
        & 62.3 
        & 58.4
        & 53.4 \\

        \rowcolor{gearRow}
        MeRF + \ourname 
        & 44.7 \gain{5.2}
        & 69.8 \gain{7.5}
        & 62.1 \gain{3.7}
        & 58.9 \gain{5.5} \\

        RuscaRL 
        & 56.9 
        & 76.7 
        & 74.2
        & 69.3 \\

        \rowcolor{gearRow}
        RuscaRL + \ourname 
        & \textbf{63.8} \gain{6.9}
        & \textbf{85.0} \gain{8.3}
        & \textbf{78.4} \gain{4.2}
        & \textbf{75.7} \gain{6.5} \\

        \bottomrule
    \end{tabular}
    \caption{
    Main results across different benchmarks.
    Aggregation-only rows differ only in reward aggregation, while plug-in rows replace the aggregation step in each host pipeline.
    All reported scores are means over three independent runs with different random seeds.
    \textbf{Bold} marks the best result within each model block.
    Parenthesized values denote absolute gains over Flat in aggregation-only rows and over the corresponding host pipeline in plug-in rows.
    }
    \label{tab:main_results}
    \vspace{-7pt}
\end{table*}

\noindent\textbf{Linear-time marginal approximation.}
\ourname computes approximate marginals $\hat q_i$ in topological order. Root
criteria use their local scores, $\hat q_i=p_i$. For a non-root criterion, the
exact marginal can be written as
\begin{equation}
    q_i
    =
    p_i\,
    \mathbb{E}_{P_{\mathcal{G}_x}(\mathbf C_{\Pa(i)}\mid\mathbf p)}
    \left[
        \prod_{j\in\Pa(i)}
        \lambda_{ji}^{1-C_j}
    \right].
    \label{eq:exact_parent_moment}
\end{equation}
We approximate this parent joint moment by a product of marginal moments. Using
$\hat q_j\approx P_{\mathcal{G}_x}(C_j=1\mid\mathbf p)$ gives the topological
update
\begin{equation}
    \hat q_i
    =
    p_i
    \prod_{j\in\Pa(i)}
    \Big(
        \hat q_j + (1-\hat q_j)\lambda_{ji}
    \Big).
    \label{eq:approx_inference}
\end{equation}
This update is exact for nodes with at most one parent, and for multi-parent
nodes whose parents are mutually independent under the conditional model;
otherwise it is a mean-field approximation to the parent joint moment. Each
parent factor lies in $[\lambda_{ji},1]$, approaching $1$ when the parent is
likely active and $\lambda_{ji}$ when it is unlikely active. Hence the update
smoothly suppresses unsupported downstream events and satisfies
\begin{equation}
    p_i
    \prod_{j\in\Pa(i)}
    \lambda_{ji}
    \le
    \hat q_i
    \le
    p_i .
    \label{eq:approx_bounds}
\end{equation}
We evaluate exact-inference agreement on small graphs in
Appendix~\ref{app:exact-inference-agreement}.

\noindent\textbf{Expected-utility reward.}
Using the approximate marginals, \ourname defines the scalar reward as
normalized expected signed utility:
\begin{equation}
    S_{\mathrm{\ourname}}(x,y)
    =
    \frac{
        \sum_{i=1}^{K} w_i \hat q_i
    }{
        \sum_{i:w_i>0} w_i
    }.
    \label{eq:gear_reward}
\end{equation}
Because the utility is additive over criteria, only marginal event probabilities
are required. Unlike flat aggregation in Eq.~\ref{eq:flat_reward}, each utility
term is weighted by a dependency-adjusted event probability that incorporates
both the local judge score and the rubric graph. The resulting scalar reward can
be used by response-level policy optimization algorithms such as PPO and GRPO
without changing the outer RL algorithm.

\noindent\textbf{Limiting cases.}
\ourname recovers simpler aggregation rules in two limits. If all retention factors
are one, then $\hat q_i=p_i$ for every criterion and Eq.~\ref{eq:gear_reward}
reduces to flat aggregation. If selected retention factors are zero and parent
decisions are binary, the corresponding child criteria are deterministically
gated by their parents, recovering hard dependency gating.

\subsection{Complexity Analysis}
\label{sec:method_complexity}

For each response, the topological update in Eq.~\ref{eq:approx_inference}
visits every criterion node once and processes every dependency edge once. The
online aggregation cost is therefore $O(K+|\mathcal{E}_x|)$ time and $O(K)$
memory for storing the marginal estimates $\hat q_i$; the final utility
aggregation in Eq.~\ref{eq:gear_reward} adds only an $O(K)$ pass. This cost is
incurred after criterion-level judge scoring, which is shared by all aggregation
variants. Graph construction is performed offline once per query-specific rubric
and is fixed across candidate responses, so \ourname adds only linear-time
graphical aggregation during policy optimization.

\section{Experiments}
\label{sec:experiments}

\subsection{Experimental Setup}
\label{sec:exp_setup}

\paragraph{Models and training.}
We use Qwen2.5-7B-Instruct
\citep{qwen2025qwen25technicalreport} and Llama-3.1-8B-Instruct
\citep{grattafiori2024llama3herdmodels} as policy models, and optimize them
with GRPO \citep{shao2024deepseekmathpushinglimitsmathematical}.
All compared aggregation methods share the same policy model, training data,
rubrics, rollout budget, and optimization pipeline; they differ only in scalar
reward aggregation.
We use Qwen3-8B for criterion-level training rewards and a separate Qwen3-32B
judge for final validation \citep{yang2025qwen3technicalreport}.
For each policy backbone and method, we run three independent training runs.
Full experimental and judge settings are provided in
Appendix~\ref{app:experimental-protocol}.

\paragraph{Benchmarks.}
We evaluate on three open-ended rubric-based benchmarks spanning different
domains: HealthBench-500 for medical dialogue quality with physician-written
rubrics \citep{arora2025healthbenchevaluatinglargelanguage}, WritingBench for
long-form writing with query-specific criteria
\citep{wu2025writingbenchcomprehensivebenchmarkgenerative}, and PLawBench for
case-based legal reasoning with problem-specific scoring rubrics
\citep{shi2026plawbenchrubricbasedbenchmarkevaluating}. 

\paragraph{Baselines.}
We evaluate \ourname in two settings. 
First, in an aggregation-only comparison, we replace only the scalarization rule
within the same minimal rubric-RL pipeline and compare against
\textbf{Flat}, which sums criterion scores as in Eq.~\ref{eq:flat_reward}, and
\textbf{Hard}, which uses the same dependency graph but replaces soft retention
with deterministic gating. 
Second, we test \ourname as a plug-in replacement for the aggregation step in
existing rubric-guided RL pipelines, including \textbf{RaR-Explicit}
\citep{gunjal2025rubricsrewardsreinforcementlearning}, \textbf{MeRF}
\citep{zhang2026simplemotivationenhancereinforcement}, and \textbf{RuscaRL}
\citep{zhou2026breakingexplorationbottleneckrubricscaffolded}.

\subsection{Main Results}
\label{sec:exp_main}

Table~\ref{tab:main_results} shows that \ourname consistently improves
rubric-based reward aggregation across benchmarks and backbones. In the
aggregation-only setting, \ourname outperforms both Flat and Hard on every
benchmark, raising the average score from $66.4$ to $72.6$ for
Qwen2.5-7B-Instruct and from $66.6$ to $73.9$ for Llama-3.1-8B-Instruct.
Hard underperforms Flat in both blocks, suggesting that deterministic gating can
discard valid partial credit under noisy criterion judgments.

As a plug-in module, \ourname also improves RaR-Explicit, MeRF, and RuscaRL on
both backbones. The best results are achieved by RuscaRL+\ourname, reaching
average scores of $74.7$ on Qwen2.5-7B-Instruct and $75.7$ on
Llama-3.1-8B-Instruct. This suggests that FCP is an aggregation-layer issue
rather than an artifact of a particular training setup.

\begin{table}[t]
    \centering
    \small
    \setlength{\tabcolsep}{6pt}
    \begin{tabular*}{\columnwidth}{@{\extracolsep{\fill}}lcc@{}}
        \toprule
        Method 
        & $L_{\mathrm{FCP}}\downarrow$ 
        & $P\uparrow$ \\
        \midrule
        Flat aggregation
        & 0.0922
        & 1.0000 \\
        Hard gating
        & 0.0000
        & 0.1877 \\
        \ourname
        & 0.0032
        & 0.2706 \\
        \bottomrule
    \end{tabular*}
    \caption{
    FCP diagnostics on edge-level dependency cases. Lower $L_{\mathrm{FCP}}$
    indicates less invalid downstream utility under violated dependencies,
    while higher $P$ indicates better preservation under satisfied dependencies.
    }
    \label{tab:fcp_diagnostics}
    \vspace{-7pt}
\end{table}

\subsection{False Credit Propagation Diagnostics}
\label{sec:exp_fcp}

We further evaluate whether \ourname reduces False Credit Propagation (FCP) at
the criterion level. The diagnostic measures a leakage--preservation trade-off:
an aggregation rule should suppress child utility when a dependency is violated.

Let $s_i^m(x,y)$ be the effective post-dependency score for criterion $i$ under
method $m$: $s_i^{\mathrm{flat}}=p_i$,
$s_i^{\mathrm{hard}}$ is the hard-gated score, and
$s_i^{\mathrm{gear}}=\hat q_i$. We define leakage cases
$\mathcal D$ as edges $(j,i)\in\mathcal E_x$ where the child is locally
supported but the parent is not, i.e., $p_i(x,y)\ge 0.5$ and
$p_j(x,y)<0.5$. The normalized leaked utility is
\begin{equation}
L_{\mathrm{FCP}}^m
=
\frac{1}{|\mathcal D|}
\sum_{(x,y,j,i)\in\mathcal D}
\frac{|w_i|}
{\sum_{k:w_k>0}w_k}
s_i^m(x,y).
\label{eq:fcp_leakage}
\end{equation}
Lower values indicate less unsupported downstream utility, including both
invalid credit and invalid penalties.

To avoid rewarding methods that simply suppress all dependent criteria, we also
measure preservation on satisfied dependencies. Let $\mathcal S$ contain edges
where both parent and child are locally supported:
$p_i(x,y)\ge 0.5$ and $p_j(x,y)\ge 0.5$. The preservation ratio is
\begin{equation}
P^m
=
\frac{1}{|\mathcal S|}
\sum_{(x,y,j,i)\in\mathcal S}
\frac{s_i^m(x,y)}{p_i(x,y)}.
\label{eq:fcp_preservation}
\end{equation}
Higher values indicate better preservation of dependency-licensed child utility.

Table~\ref{tab:fcp_diagnostics} shows that \ourname reduces
$L_{\mathrm{FCP}}$ from $0.0922$ to $0.0032$, a $96.5\%$ relative reduction
over Flat. Hard gating achieves slightly lower leakage, but preserves much less
licensed utility than \ourname ($0.1877$ vs. $0.2706$). Thus, \ourname keeps
leakage close to zero while avoiding the over-suppression caused by hard gating. Additional details on diagnostic case construction and interpretation
are provided in Appendix~\ref{app:fcp-diagnostics}.

\begin{table}[t]
    \centering
    \small
    \setlength{\tabcolsep}{6pt}
    \renewcommand{\arraystretch}{1.12}    
    \begin{tabular}{@{}lc@{}}
        \toprule
        Statistic & HealthBench \\
        \midrule
        Num. rubrics & 5000 \\
        Avg. criteria / rubric & 11.48 \\
        Avg. edges / rubric & 10.92 \\
        Avg. update size $K+|\mathcal E_x|$ & 22.40 \\
        Non-empty graph rate & 91.84\% \\
        Invalid edge candidates / rubric & 0.42 \\
        Online aggregation latency / response & 4.1 ms \\
        \bottomrule
    \end{tabular}
    \caption{
    Graph construction and online aggregation statistics on HealthBench, the largest rubric benchmark in our experiments.
    Latency measures only post-judge aggregation.
    }
    \label{tab:efficiency}
    \vspace{-5pt}
\end{table}

\subsection{Efficiency and Construction Statistics}
\label{sec:exp_efficiency}

We evaluate the overhead of dependency-structured aggregation. Graph construction
is performed offline once per query-specific rubric and fixed across candidate
responses. At reward time, \ourname applies the topological update in
Eq.~\ref{eq:approx_inference}, requiring $O(K+|\mathcal E_x|)$ time and $O(K)$
memory after local criterion scores are available.

Table~\ref{tab:efficiency} reports results on HealthBench. The average rubric
has $11.48$ criteria and $10.92$ edges, with an average update size of $22.40$.
Online aggregation takes 4.1 ms per response, excluding generation and
criterion-level judging. After validation and DAG projection, $91.84\%$ of
rubrics yield non-empty graphs, with $0.42$ invalid edge candidates per rubric.
Full statistics across all benchmarks are reported in Appendix~\ref{app:efficiency-full}.

\begin{figure}[t]
    \centering
    \includegraphics[width=\columnwidth]{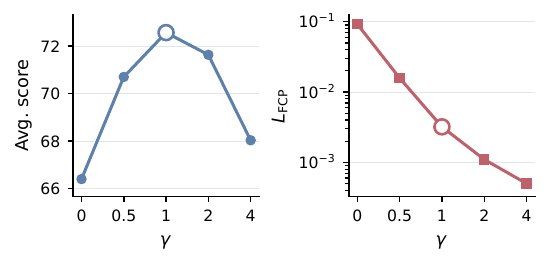}
    \caption{
    Sensitivity to the suppression-strength parameter $\gamma$.
    Larger $\gamma$ imposes stronger dependency suppression; hollow points mark the
    default setting.
    }
    \label{fig:lambda_sensitivity}
    \vspace{-6pt}
\end{figure}

\begin{table}[t]
    \centering
    \small
    \setlength{\tabcolsep}{6pt}
    \renewcommand{\arraystretch}{1.12}
    \begin{tabular}{@{}lcc@{}}
        \toprule
        \textbf{Variant}
        & \textbf{Score} $\uparrow$
        & \textbf{$\Delta$ vs. Flat} \\
        \midrule
        No graph / Flat
        & 56.7
        & -- \\
        Random graph
        & 55.8
        & $-0.9$ \\
        Untyped graph
        & 60.4
        & $+3.7$ \\
        Full graph + hard gating
        & 54.6
        & $-2.1$ \\
        \rowcolor{gearRow}
        Full graph + soft inference (\ourname)
        & \textbf{62.9}
        & \textbf{$+6.2$} \\
        \bottomrule
    \end{tabular}
    \caption{ 
    Ablations on HealthBench-500. All variants
    differ only in the aggregation module; full results are reported in Appendix~\ref{app:ablations}.
    }
    \label{tab:ablations}
    \vspace{-6pt}
\end{table}

\subsection{Sensitivity Analysis}
\label{sec:exp_sensitivity}

We analyze sensitivity to the global suppression parameter $\gamma$.
With the rubric graph and edge types fixed, each type-level retention factor is
\begin{equation}
    \lambda_t(\gamma)=
    \begin{cases}
        1, & \gamma=0,\\
        \lambda_t^{\gamma}, & \gamma>0,
    \end{cases}
    \label{eq:gamma_sensitivity}
\end{equation}
where $t\in\{\mathrm{wk},\mathrm{st},\mathrm{act}\}$ and $\lambda_t$ is the
default retention factor. Here, $\gamma=0$ recovers Flat aggregation,
$\gamma=1$ gives the default \ourname, and larger values impose stronger
suppression.

Figure~\ref{fig:lambda_sensitivity} shows that increasing $\gamma$ sharply
reduces FCP leakage. The default setting lowers $L_{\mathrm{FCP}}$ from
$0.0922$ to $0.0032$ while achieving the best average benchmark score. Stronger
suppression further reduces leakage but degrades task performance, indicating
that moderate soft suppression gives the best score--leakage trade-off.

\subsection{Ablation Studies}
\label{sec:exp_ablations}

We ablate dependency topology, relation typing, and soft graphical inference
under the same minimal rubric-RL setting as Table~\ref{tab:main_results}.
Random graph rewires edges while preserving edge counts and type frequencies;
Untyped graph uses the annotated topology with a shared retention factor; Hard
gating replaces soft retention with deterministic gating.

Table~\ref{tab:ablations} shows that random rewiring fails to improve over Flat,
whereas the untyped graph improves from $56.7$ to $60.4$.
The full typed graph with soft inference performs best at $62.9$, while hard
gating underperforms Flat. These results support meaningful topology, typed
dependencies, and soft inference. Full three-benchmark ablations are reported
in Appendix~\ref{app:ablations}.

\subsection{Additional Validation and Robustness}
\label{sec:additional-validation}

\paragraph{Multi-parent aggregation.}
GEAR uses a product rule to combine support from multiple parents.
Among dependent criteria, 70.0\% have one parent, 24.4\% have two
parents, and only 5.6\% have three or more, with an average of 1.37
parents.
We additionally compare the product rule with geometric-mean,
arithmetic-mean, and maximum-support aggregation under the same
Qwen2.5-7B aggregation-only setup.
The product rule achieves the best average score and leakage
(\(72.6/0.0032\)), compared with the geometric mean
(\(70.5/0.0058\)), arithmetic mean (\(69.4/0.0087\)), and maximum
support (\(68.8/0.0146\)).
All four dependency-aware variants outperform Flat (66.4), indicating
that the gains are not unique to the product formulation, while the
product rule provides the strongest score--leakage trade-off.

\paragraph{Rubric specification versus aggregation.}
GEAR distinguishes criterion omission, relational under-specification,
and aggregation failure.
It cannot recover an omitted criterion, but it can make an implicit
licensing relation explicit and incorporate that relation into reward
aggregation.
To test whether textualizing the dependencies is sufficient, we construct
a \textsc{Conditional Rewrite} baseline that inserts the same prerequisite
and activation conditions into each child criterion before applying Flat
aggregation.
All variants use the same dependency annotations, criterion weights,
judges, training budgets, and random seeds.

\begin{table}[t]
    \centering
    \small
    \setlength{\tabcolsep}{3.5pt}
    \resizebox{\columnwidth}{!}{
    \begin{tabular}{lccc}
        \toprule
        Method & Qwen2.5-7B & Llama-3.1-8B & Avg. \\
        \midrule
        Original + Flat
            & 66.4 & 66.6 & 66.5 \\
        Conditional Rewrite + Flat
            & 69.9 & 70.4 & 70.2 \\
        Original + GEAR
            & \textbf{72.6} & \textbf{73.9} & \textbf{73.3} \\
        \bottomrule
    \end{tabular}
    }
    \caption{Comparison with dependency-aware rubric rewriting.
    Results are averaged across the three benchmarks.}
    \label{tab:conditional-rewrite}
    \vspace{-5pt}
\end{table}

As shown in Table~\ref{tab:conditional-rewrite}, conditional rewriting
recovers part of the gap, confirming that rubric specification matters.
GEAR provides a further improvement by retaining decomposed parent and
child judgments and explicitly aggregating their dependency support.

\paragraph{Graph quality and human validation.}
We manually audit 180 retained dependency edges, with 60 edges sampled
from each benchmark and all relation types represented.
Two annotators independently label each edge before adjudication.
The audit yields 91.7\% valid-edge precision, 87.3\% relation-type
accuracy among valid edges, and a pooled Cohen's \(\kappa\) of 0.80.
This audit evaluates the precision of retained edges rather than the
recall of omitted dependencies.

We further annotate 300 dependency cases using human labels determined
before inspecting method outputs.
On these human-defined subsets, GEAR reduces violated-dependency leakage
from 0.082 for Flat to 0.010, while preserving 0.648 licensed downstream
utility, compared with 0.231 for Hard gating.
This human-grounded audit complements the scalable mechanism-level
diagnostic in Section~\ref{sec:exp_fcp}.
Finally, in a blind evaluation of 250 matched Flat--GEAR response pairs,
GEAR is preferred in 72.9\% of non-tied comparisons
(95\% Wilson CI: 66.7--78.3\%), with inter-annotator agreement
\(\kappa=0.72\).

\paragraph{Independent and cross-family evaluation.}
We re-evaluate the aggregation-only policies using benchmark-specific
protocols: the official HealthBench setup with GPT-4.1, the released
WritingBench critic, and the released PLawBench evaluation prompt with
Gemini.
As shown in Table~\ref{tab:benchmark-specific-evaluation}, GEAR
consistently outperforms Flat and Hard, with gains of 5.4--6.3 points
over Flat.

\begin{table}[t]
    \centering
    \small
    \setlength{\tabcolsep}{4.5pt}
    \begin{tabular}{lccc}
        \toprule
        Benchmark & Flat & Hard & GEAR \\
        \midrule
        HealthBench
            & 55.2 & 53.6 & \textbf{61.5} \\
        WritingBench
            & 74.4 & 72.0 & \textbf{79.8} \\
        PLawBench
            & 68.1 & 65.5 & \textbf{74.4} \\
        \bottomrule
    \end{tabular}
    \caption{Evaluation with benchmark-specific protocols.
    Results are averaged over both policy backbones and three seeds.}
    \label{tab:benchmark-specific-evaluation}
    \vspace{-5pt}
\end{table}

To test model-family dependence, we also jointly replace the graph
builder, training-time reward judge, and final evaluator with non-Qwen
models while keeping the policy, training protocol, and random seeds
fixed.
GEAR retains a 5.9-point improvement over its matched Flat baseline,
suggesting that the observed gains are not specific to a Qwen-centric
annotation and evaluation pipeline.

\section{Conclusion}
\label{sec:conclusion}

We introduced False Credit Propagation (FCP), a structural failure in flat
rubric aggregation where criteria contribute reward or penalty without
dependencies licensing their utility. To address FCP, we proposed \ourname, a
graphical aggregation framework that models criteria as latent events and uses
typed prerequisite and activation relations to infer dependency-adjusted
marginals. This yields linear-time reward computation for rubric-based
reinforcement learning pipelines without changing the outer optimization
algorithm.
Across benchmarks, backbones, and rubric-guided RL pipelines, \ourname improves
task performance over flat aggregation and hard gating while reducing invalid
downstream utility. The results suggest that the aggregation layer itself is an
important part of reward-model design, especially when rubric criteria are
semantically dependent. Overall, rubric dependencies are not merely annotation
structure: they should be part of the reward model, not flattened into
independent checklist items.

\section{Limitations}
\label{sec:limitations}

\ourname focuses on reward aggregation given criterion-level scores and a
dependency-structured rubric.
It models monotone prerequisite and activation relations, where an
unsupported parent can only suppress a child's contribution.
It does not cover compensatory, alternative, mutually exclusive,
non-monotonic, or cyclic relations, nor can it recover criteria omitted
from the rubric.
Rubric generation and dependency-aware aggregation therefore remain
complementary.

The quality of the resulting reward still depends on the local criterion
judgments and the constructed graph.
Incorrect or missing edges may suppress valid utility or leave false
credit uncorrected.
Our human audit evaluates the precision and type accuracy of retained
edges, but does not measure the recall of omitted dependencies.
Likewise, the scalable FCP diagnostic is based on local judge scores;
the human-grounded audit provides complementary evidence but covers only
a limited subset of cases.

\ourname uses fixed type-level retention factors and bounded soft judge
scores, which need not be calibrated probabilities.
These choices make the method simple and reproducible, but may not capture
fine-grained variation across tasks or edges.
In addition, the topological update becomes a mean-field approximation
when multiple parent events are correlated.
Learning edge-specific retention factors and using more accurate inference
for dense subgraphs are promising extensions.

Finally, our experiments cover three open-ended domains, multiple policy
backbones, independent evaluators, and human preference judgments, but do
not exhaust possible rubric structures or deployment settings.
The human evaluation measures perceived rubric alignment rather than
domain-expert factual correctness.
Broader expert validation and richer dependency types remain important
directions for future work.

\section*{Acknowledgments}

This work is supported by the National Science Foundation of China (62401327, 62672026), and the Beijing Major Science and Technology Project under Contract no. Z25110008125031. This work  was supported  by Beijing Academy  of Artificial Intelligence (BAAI).

\bibliography{custom}

@misc{ouyang2022traininglanguagemodelsfollow,
      title={Training language models to follow instructions with human feedback}, 
      author={Long Ouyang and Jeff Wu and Xu Jiang and Diogo Almeida and Carroll L. Wainwright and Pamela Mishkin and Chong Zhang and Sandhini Agarwal and Katarina Slama and Alex Ray and John Schulman and Jacob Hilton and Fraser Kelton and Luke Miller and Maddie Simens and Amanda Askell and Peter Welinder and Paul Christiano and Jan Leike and Ryan Lowe},
      year={2022},
      eprint={2203.02155},
      archivePrefix={arXiv},
      primaryClass={cs.CL},
      url={https://arxiv.org/abs/2203.02155}, 
}

@misc{bai2022constitutionalaiharmlessnessai,
      title={Constitutional AI: Harmlessness from AI Feedback}, 
      author={Yuntao Bai and Saurav Kadavath and Sandipan Kundu and Amanda Askell and Jackson Kernion and Andy Jones and Anna Chen and Anna Goldie and Azalia Mirhoseini and Cameron McKinnon and Carol Chen and Catherine Olsson and Christopher Olah and Danny Hernandez and Dawn Drain and Deep Ganguli and Dustin Li and Eli Tran-Johnson and Ethan Perez and Jamie Kerr and Jared Mueller and Jeffrey Ladish and Joshua Landau and Kamal Ndousse and Kamile Lukosuite and Liane Lovitt and Michael Sellitto and Nelson Elhage and Nicholas Schiefer and Noemi Mercado and Nova DasSarma and Robert Lasenby and Robin Larson and Sam Ringer and Scott Johnston and Shauna Kravec and Sheer El Showk and Stanislav Fort and Tamera Lanham and Timothy Telleen-Lawton and Tom Conerly and Tom Henighan and Tristan Hume and Samuel R. Bowman and Zac Hatfield-Dodds and Ben Mann and Dario Amodei and Nicholas Joseph and Sam McCandlish and Tom Brown and Jared Kaplan},
      year={2022},
      eprint={2212.08073},
      archivePrefix={arXiv},
      primaryClass={cs.CL},
      url={https://arxiv.org/abs/2212.08073}, 
}

@misc{rafailov2024directpreferenceoptimizationlanguage,
      title={Direct Preference Optimization: Your Language Model is Secretly a Reward Model}, 
      author={Rafael Rafailov and Archit Sharma and Eric Mitchell and Stefano Ermon and Christopher D. Manning and Chelsea Finn},
      year={2024},
      eprint={2305.18290},
      archivePrefix={arXiv},
      primaryClass={cs.LG},
      url={https://arxiv.org/abs/2305.18290}, 
}

@misc{jin2025multidimensionalrubricorientedrewardmodel,
      title={Multidimensional Rubric-oriented Reward Model Learning via Geometric Projection Reference Constraints}, 
      author={Yongnan Jin and Xurui Li and Feng Cao and Liucun Gao and Juanjuan Yao},
      year={2025},
      eprint={2511.16139},
      archivePrefix={arXiv},
      primaryClass={cs.AI},
      url={https://arxiv.org/abs/2511.16139}, 
}

@misc{arora2025healthbenchevaluatinglargelanguage,
      title={HealthBench: Evaluating Large Language Models Towards Improved Human Health}, 
      author={Rahul K. Arora and Jason Wei and Rebecca Soskin Hicks and Preston Bowman and Joaquin Quiñonero-Candela and Foivos Tsimpourlas and Michael Sharman and Meghan Shah and Andrea Vallone and Alex Beutel and Johannes Heidecke and Karan Singhal},
      year={2025},
      eprint={2505.08775},
      archivePrefix={arXiv},
      primaryClass={cs.CL},
      url={https://arxiv.org/abs/2505.08775}, 
}

@misc{rein2023gpqagraduatelevelgoogleproofqa,
      title={GPQA: A Graduate-Level Google-Proof Q\&A Benchmark}, 
      author={David Rein and Betty Li Hou and Asa Cooper Stickland and Jackson Petty and Richard Yuanzhe Pang and Julien Dirani and Julian Michael and Samuel R. Bowman},
      year={2023},
      eprint={2311.12022},
      archivePrefix={arXiv},
      primaryClass={cs.AI},
      url={https://arxiv.org/abs/2311.12022}, 
}

@misc{he2025advancedifrubricbasedbenchmarkingreinforcement,
      title={AdvancedIF: Rubric-Based Benchmarking and Reinforcement Learning for Advancing LLM Instruction Following}, 
      author={Yun He and Wenzhe Li and Hejia Zhang and Songlin Li and Karishma Mandyam and Sopan Khosla and Yuanhao Xiong and Nanshu Wang and Xiaoliang Peng and Beibin Li and Shengjie Bi and Shishir G. Patil and Qi Qi and Shengyu Feng and Julian Katz-Samuels and Richard Yuanzhe Pang and Sujan Gonugondla and Hunter Lang and Yue Yu and Yundi Qian and Maryam Fazel-Zarandi and Licheng Yu and Amine Benhalloum and Hany Awadalla and Manaal Faruqui},
      year={2025},
      eprint={2511.10507},
      archivePrefix={arXiv},
      primaryClass={cs.CL},
      url={https://arxiv.org/abs/2511.10507}, 
}

@misc{li2026rubrichubcomprehensivehighlydiscriminative,
      title={RubricHub: A Comprehensive and Highly Discriminative Rubric Dataset via Automated Coarse-to-Fine Generation}, 
      author={Sunzhu Li and Jiale Zhao and Miteto Wei and Huimin Ren and Yang Zhou and Jingwen Yang and Shunyu Liu and Kaike Zhang and Wei Chen},
      year={2026},
      eprint={2601.08430},
      archivePrefix={arXiv},
      primaryClass={cs.AI},
      url={https://arxiv.org/abs/2601.08430}, 
}

@misc{gunjal2025rubricsrewardsreinforcementlearning,
      title={Rubrics as Rewards: Reinforcement Learning Beyond Verifiable Domains}, 
      author={Anisha Gunjal and Anthony Wang and Elaine Lau and Vaskar Nath and Yunzhong He and Bing Liu and Sean Hendryx},
      year={2025},
      eprint={2507.17746},
      archivePrefix={arXiv},
      primaryClass={cs.LG},
      url={https://arxiv.org/abs/2507.17746}, 
}

@misc{hong2026rulerslockedrubricsevidenceanchored,
      title={RULERS: Locked Rubrics and Evidence-Anchored Scoring for Robust LLM Evaluation}, 
      author={Yihan Hong and Huaiyuan Yao and Bolin Shen and Wanpeng Xu and Hua Wei and Yushun Dong},
      year={2026},
      eprint={2601.08654},
      archivePrefix={arXiv},
      primaryClass={cs.CL},
      url={https://arxiv.org/abs/2601.08654}, 
}

@misc{pan2026rubricevalrubriclevelmetaevaluationbenchmark,
      title={RubricEval: A Rubric-Level Meta-Evaluation Benchmark for LLM Judges in Instruction Following}, 
      author={Tianjun Pan and Xuan Lin and Wenyan Yang and Qianyu He and Shisong Chen and Licai Qi and Wanqing Xu and Hongwei Feng and Bo Xu and Yanghua Xiao},
      year={2026},
      eprint={2603.25133},
      archivePrefix={arXiv},
      primaryClass={cs.AI},
      url={https://arxiv.org/abs/2603.25133}, 
}

@inproceedings{NEURIPS2023_91f18a12,
 author = {Zheng, Lianmin and Chiang, Wei-Lin and Sheng, Ying and Zhuang, Siyuan and Wu, Zhanghao and Zhuang, Yonghao and Lin, Zi and Li, Zhuohan and Li, Dacheng and Xing, Eric and Zhang, Hao and Gonzalez, Joseph and Stoica, Ion},
 booktitle = {Advances in Neural Information Processing Systems},
 editor = {A. Oh and T. Naumann and A. Globerson and K. Saenko and M. Hardt and S. Levine},
 pages = {46595--46623},
 publisher = {Curran Associates, Inc.},
 title = {Judging LLM-as-a-Judge with MT-Bench and Chatbot Arena},
 url = {https://proceedings.neurips.cc/paper_files/paper/2023/file/91f18a1287b398d378ef22505bf41832-Paper-Datasets_and_Benchmarks.pdf},
 volume = {36},
 year = {2023}
}

@misc{lambert2024rewardbenchevaluatingrewardmodels,
      title={RewardBench: Evaluating Reward Models for Language Modeling}, 
      author={Nathan Lambert and Valentina Pyatkin and Jacob Morrison and LJ Miranda and Bill Yuchen Lin and Khyathi Chandu and Nouha Dziri and Sachin Kumar and Tom Zick and Yejin Choi and Noah A. Smith and Hannaneh Hajishirzi},
      year={2024},
      eprint={2403.13787},
      archivePrefix={arXiv},
      primaryClass={cs.LG},
      url={https://arxiv.org/abs/2403.13787}, 
}

@book{pearl2014probabilistic,
  title={Probabilistic reasoning in intelligent systems: networks of plausible inference},
  author={Pearl, Judea},
  year={2014},
  publisher={Elsevier}
}

@book{koller2009probabilistic,
  title={Probabilistic graphical models: principles and techniques},
  author={Koller, Daphne and Friedman, Nir},
  year={2009},
  publisher={MIT press}
}

@misc{liu2023gevalnlgevaluationusing,
      title={G-Eval: NLG Evaluation using GPT-4 with Better Human Alignment}, 
      author={Yang Liu and Dan Iter and Yichong Xu and Shuohang Wang and Ruochen Xu and Chenguang Zhu},
      year={2023},
      eprint={2303.16634},
      archivePrefix={arXiv},
      primaryClass={cs.CL},
      url={https://arxiv.org/abs/2303.16634}, 
}

@misc{kim2024prometheusinducingfinegrainedevaluation,
      title={Prometheus: Inducing Fine-grained Evaluation Capability in Language Models}, 
      author={Seungone Kim and Jamin Shin and Yejin Cho and Joel Jang and Shayne Longpre and Hwaran Lee and Sangdoo Yun and Seongjin Shin and Sungdong Kim and James Thorne and Minjoon Seo},
      year={2024},
      eprint={2310.08491},
      archivePrefix={arXiv},
      primaryClass={cs.CL},
      url={https://arxiv.org/abs/2310.08491}, 
}

@misc{kim2024prometheus2opensource,
      title={Prometheus 2: An Open Source Language Model Specialized in Evaluating Other Language Models}, 
      author={Seungone Kim and Juyoung Suk and Shayne Longpre and Bill Yuchen Lin and Jamin Shin and Sean Welleck and Graham Neubig and Moontae Lee and Kyungjae Lee and Minjoon Seo},
      year={2024},
      eprint={2405.01535},
      archivePrefix={arXiv},
      primaryClass={cs.CL},
      url={https://arxiv.org/abs/2405.01535}, 
}

@misc{gu2025surveyllmasajudge,
      title={A Survey on LLM-as-a-Judge}, 
      author={Jiawei Gu and Xuhui Jiang and Zhichao Shi and Hexiang Tan and Xuehao Zhai and Chengjin Xu and Wei Li and Yinghan Shen and Shengjie Ma and Honghao Liu and Saizhuo Wang and Kun Zhang and Yuanzhuo Wang and Wen Gao and Lionel Ni and Jian Guo},
      year={2025},
      eprint={2411.15594},
      archivePrefix={arXiv},
      primaryClass={cs.CL},
      url={https://arxiv.org/abs/2411.15594}, 
}

@misc{wu2025writingbenchcomprehensivebenchmarkgenerative,
      title={WritingBench: A Comprehensive Benchmark for Generative Writing}, 
      author={Yuning Wu and Jiahao Mei and Ming Yan and Chenliang Li and Shaopeng Lai and Yuran Ren and Zijia Wang and Ji Zhang and Mengyue Wu and Qin Jin and Fei Huang},
      year={2025},
      eprint={2503.05244},
      archivePrefix={arXiv},
      primaryClass={cs.AI},
      url={https://arxiv.org/abs/2503.05244}, 
}

@inproceedings{galvan-sosa-etal-2025-rubriks,
    title = "Rubrik{'}s Cube: Testing a New Rubric for Evaluating Explanations on the {CUBE} dataset",
    author = "Galvan-Sosa, Diana  and
      Gaudeau, Gabrielle  and
      Kavumba, Pride  and
      Li, Yunmeng  and
      Gu, Hongyi  and
      Yuan, Zheng  and
      Sakaguchi, Keisuke  and
      Buttery, Paula",
    editor = "Che, Wanxiang  and
      Nabende, Joyce  and
      Shutova, Ekaterina  and
      Pilehvar, Mohammad Taher",
    booktitle = "Proceedings of the 63rd Annual Meeting of the Association for Computational Linguistics (Volume 1: Long Papers)",
    month = jul,
    year = "2025",
    address = "Vienna, Austria",
    publisher = "Association for Computational Linguistics",
    url = "https://aclanthology.org/2025.acl-long.1160/",
    doi = "10.18653/v1/2025.acl-long.1160",
    pages = "23800--23839",
    ISBN = "979-8-89176-251-0"
}

@misc{sharma2025researchrubricsbenchmarkpromptsrubrics,
      title={ResearchRubrics: A Benchmark of Prompts and Rubrics For Evaluating Deep Research Agents}, 
      author={Manasi Sharma and Chen Bo Calvin Zhang and Chaithanya Bandi and Clinton Wang and Ankit Aich and Huy Nghiem and Tahseen Rabbani and Ye Htet and Brian Jang and Sumana Basu and Aishwarya Balwani and Denis Peskoff and Marcos Ayestaran and Sean M. Hendryx and Brad Kenstler and Bing Liu},
      year={2025},
      eprint={2511.07685},
      archivePrefix={arXiv},
      primaryClass={cs.AI},
      url={https://arxiv.org/abs/2511.07685}, 
}

@misc{han2026deerbenchmarkevaluatingdeep,
      title={DEER: A Benchmark for Evaluating Deep Research Agents on Expert Report Generation}, 
      author={Janghoon Han and Heegyu Kim and Changho Lee and Dahm Lee and Min Hyung Park and Hosung Song and Stanley Jungkyu Choi and Moontae Lee and Honglak Lee},
      year={2026},
      eprint={2512.17776},
      archivePrefix={arXiv},
      primaryClass={cs.CL},
      url={https://arxiv.org/abs/2512.17776}, 
}

@misc{huang2025reinforcementlearningrubricanchors,
      title={Reinforcement Learning with Rubric Anchors}, 
      author={Zenan Huang and Yihong Zhuang and Guoshan Lu and Zeyu Qin and Haokai Xu and Tianyu Zhao and Ru Peng and Jiaqi Hu and Zhanming Shen and Xiaomeng Hu and Xijun Gu and Peiyi Tu and Jiaxin Liu and Wenyu Chen and Yuzhuo Fu and Zhiting Fan and Yanmei Gu and Yuanyuan Wang and Zhengkai Yang and Jianguo Li and Junbo Zhao},
      year={2025},
      eprint={2508.12790},
      archivePrefix={arXiv},
      primaryClass={cs.AI},
      url={https://arxiv.org/abs/2508.12790}, 
}

@misc{zhou2026breakingexplorationbottleneckrubricscaffolded,
      title={Breaking the Exploration Bottleneck: Rubric-Scaffolded Reinforcement Learning for General LLM Reasoning}, 
      author={Yang Zhou and Sunzhu Li and Shunyu Liu and Wenkai Fang and Kongcheng Zhang and Jiale Zhao and Jingwen Yang and Yihe Zhou and Jianwei Lv and Tongya Zheng and Hengtong Lu and Wei Chen and Yan Xie and Mingli Song},
      year={2026},
      eprint={2508.16949},
      archivePrefix={arXiv},
      primaryClass={cs.LG},
      url={https://arxiv.org/abs/2508.16949}, 
}

@misc{zhang2026simplemotivationenhancereinforcement,
      title={A Simple "Motivation" Can Enhance Reinforcement Finetuning of Large Reasoning Models}, 
      author={Junjie Zhang and Guozheng Ma and Shunyu Liu and Haoyu Wang and Jiaxing Huang and Ting-En Lin and Fei Huang and Yongbin Li and Dacheng Tao},
      year={2026},
      eprint={2506.18485},
      archivePrefix={arXiv},
      primaryClass={cs.CL},
      url={https://arxiv.org/abs/2506.18485}, 
}

@InProceedings{10.1007/978-3-031-11644-5_29,
author="Condor, Aubrey
and Pardos, Zachary
and Linn, Marcia",
editor="Rodrigo, Maria Mercedes
and Matsuda, Noburu
and Cristea, Alexandra I.
and Dimitrova, Vania",
title="Representing Scoring Rubrics as Graphs for Automatic Short Answer Grading",
booktitle="Artificial Intelligence  in Education",
year="2022",
publisher="Springer International Publishing",
address="Cham",
pages="354--365",
isbn="978-3-031-11644-5"
}

@misc{starace2025paperbenchevaluatingaisability,
      title={PaperBench: Evaluating AI's Ability to Replicate AI Research}, 
      author={Giulio Starace and Oliver Jaffe and Dane Sherburn and James Aung and Jun Shern Chan and Leon Maksin and Rachel Dias and Evan Mays and Benjamin Kinsella and Wyatt Thompson and Johannes Heidecke and Amelia Glaese and Tejal Patwardhan},
      year={2025},
      eprint={2504.01848},
      archivePrefix={arXiv},
      primaryClass={cs.AI},
      url={https://arxiv.org/abs/2504.01848}, 
}

@misc{shi2026plawbenchrubricbasedbenchmarkevaluating,
      title={PLawBench: A Rubric-Based Benchmark for Evaluating LLMs in Real-World Legal Practice}, 
      author={Yuzhen Shi and Huanghai Liu and Yiran Hu and Gaojie Song and Xinran Xu and Yubo Ma and Tianyi Tang and Li Zhang and Qingjing Chen and Di Feng and Wenbo Lv and Weiheng Wu and Kexin Yang and Sen Yang and Wei Wang and Rongyao Shi and Yuanyang Qiu and Yuemeng Qi and Jingwen Zhang and Xiaoyu Sui and Yifan Chen and Yi Zhang and An Yang and Bowen Yu and Dayiheng Liu and Junyang Lin and Weixing Shen and Bing Zhao and Charles L. A. Clarke and Hu Wei},
      year={2026},
      eprint={2601.16669},
      archivePrefix={arXiv},
      primaryClass={cs.CL},
      url={https://arxiv.org/abs/2601.16669}, 
}

@misc{qwen2025qwen25technicalreport,
      title={Qwen2.5 Technical Report}, 
      author={Qwen and : and An Yang and Baosong Yang and Beichen Zhang and Binyuan Hui and Bo Zheng and Bowen Yu and Chengyuan Li and Dayiheng Liu and Fei Huang and Haoran Wei and Huan Lin and Jian Yang and Jianhong Tu and Jianwei Zhang and Jianxin Yang and Jiaxi Yang and Jingren Zhou and Junyang Lin and Kai Dang and Keming Lu and Keqin Bao and Kexin Yang and Le Yu and Mei Li and Mingfeng Xue and Pei Zhang and Qin Zhu and Rui Men and Runji Lin and Tianhao Li and Tianyi Tang and Tingyu Xia and Xingzhang Ren and Xuancheng Ren and Yang Fan and Yang Su and Yichang Zhang and Yu Wan and Yuqiong Liu and Zeyu Cui and Zhenru Zhang and Zihan Qiu},
      year={2025},
      eprint={2412.15115},
      archivePrefix={arXiv},
      primaryClass={cs.CL},
      url={https://arxiv.org/abs/2412.15115}, 
}

@misc{grattafiori2024llama3herdmodels,
      title={The Llama 3 Herd of Models}, 
      author={Aaron Grattafiori and Abhimanyu Dubey and Abhinav Jauhri and Abhinav Pandey and Abhishek Kadian and Ahmad Al-Dahle and Aiesha Letman and Akhil Mathur and Alan Schelten and Alex Vaughan and Amy Yang and Angela Fan and Anirudh Goyal and Anthony Hartshorn and Aobo Yang and Archi Mitra and Archie Sravankumar and Artem Korenev and Arthur Hinsvark and Arun Rao and Aston Zhang and Aurelien Rodriguez and Austen Gregerson and Ava Spataru and Baptiste Roziere and Bethany Biron and Binh Tang and Bobbie Chern and Charlotte Caucheteux and Chaya Nayak and Chloe Bi and Chris Marra and Chris McConnell and Christian Keller and Christophe Touret and Chunyang Wu and Corinne Wong and Cristian Canton Ferrer and Cyrus Nikolaidis and Damien Allonsius and Daniel Song and Danielle Pintz and Danny Livshits and Danny Wyatt and David Esiobu and Dhruv Choudhary and Dhruv Mahajan and Diego Garcia-Olano and Diego Perino and Dieuwke Hupkes and Egor Lakomkin and Ehab AlBadawy and Elina Lobanova and Emily Dinan and Eric Michael Smith and Filip Radenovic and Francisco Guzmán and Frank Zhang and Gabriel Synnaeve and Gabrielle Lee and Georgia Lewis Anderson and Govind Thattai and Graeme Nail and Gregoire Mialon and Guan Pang and Guillem Cucurell and Hailey Nguyen and Hannah Korevaar and Hu Xu and Hugo Touvron and Iliyan Zarov and Imanol Arrieta Ibarra and Isabel Kloumann and Ishan Misra and Ivan Evtimov and Jack Zhang and Jade Copet and Jaewon Lee and Jan Geffert and Jana Vranes and Jason Park and Jay Mahadeokar and Jeet Shah and Jelmer van der Linde and Jennifer Billock and Jenny Hong and Jenya Lee and Jeremy Fu and Jianfeng Chi and Jianyu Huang and Jiawen Liu and Jie Wang and Jiecao Yu and Joanna Bitton and Joe Spisak and Jongsoo Park and Joseph Rocca and Joshua Johnstun and Joshua Saxe and Junteng Jia and Kalyan Vasuden Alwala and Karthik Prasad and Kartikeya Upasani and Kate Plawiak and Ke Li and Kenneth Heafield and Kevin Stone and Khalid El-Arini and Krithika Iyer and Kshitiz Malik and Kuenley Chiu and Kunal Bhalla and Kushal Lakhotia and Lauren Rantala-Yeary and Laurens van der Maaten and Lawrence Chen and Liang Tan and Liz Jenkins and Louis Martin and Lovish Madaan and Lubo Malo and Lukas Blecher and Lukas Landzaat and Luke de Oliveira and Madeline Muzzi and Mahesh Pasupuleti and Mannat Singh and Manohar Paluri and Marcin Kardas and Maria Tsimpoukelli and Mathew Oldham and Mathieu Rita and Maya Pavlova and Melanie Kambadur and Mike Lewis and Min Si and Mitesh Kumar Singh and Mona Hassan and Naman Goyal and Narjes Torabi and Nikolay Bashlykov and Nikolay Bogoychev and Niladri Chatterji and Ning Zhang and Olivier Duchenne and Onur Çelebi and Patrick Alrassy and Pengchuan Zhang and Pengwei Li and Petar Vasic and Peter Weng and Prajjwal Bhargava and Pratik Dubal and Praveen Krishnan and Punit Singh Koura and Puxin Xu and Qing He and Qingxiao Dong and Ragavan Srinivasan and Raj Ganapathy and Ramon Calderer and Ricardo Silveira Cabral and Robert Stojnic and Roberta Raileanu and Rohan Maheswari and Rohit Girdhar and Rohit Patel and Romain Sauvestre and Ronnie Polidoro and Roshan Sumbaly and Ross Taylor and Ruan Silva and Rui Hou and Rui Wang and Saghar Hosseini and Sahana Chennabasappa and Sanjay Singh and Sean Bell and Seohyun Sonia Kim and Sergey Edunov and Shaoliang Nie and Sharan Narang and Sharath Raparthy and Sheng Shen and Shengye Wan and Shruti Bhosale and Shun Zhang and Simon Vandenhende and Soumya Batra and Spencer Whitman and Sten Sootla and Stephane Collot and Suchin Gururangan and Sydney Borodinsky and Tamar Herman and Tara Fowler and Tarek Sheasha and Thomas Georgiou and Thomas Scialom and Tobias Speckbacher and Todor Mihaylov and Tong Xiao and Ujjwal Karn and Vedanuj Goswami and Vibhor Gupta and Vignesh Ramanathan and Viktor Kerkez and Vincent Gonguet and Virginie Do and Vish Vogeti and Vítor Albiero and Vladan Petrovic and Weiwei Chu and Wenhan Xiong and Wenyin Fu and Whitney Meers and Xavier Martinet and Xiaodong Wang and Xiaofang Wang and Xiaoqing Ellen Tan and Xide Xia and Xinfeng Xie and Xuchao Jia and Xuewei Wang and Yaelle Goldschlag and Yashesh Gaur and Yasmine Babaei and Yi Wen and Yiwen Song and Yuchen Zhang and Yue Li and Yuning Mao and Zacharie Delpierre Coudert and Zheng Yan and Zhengxing Chen and Zoe Papakipos and Aaditya Singh and Aayushi Srivastava and Abha Jain and Adam Kelsey and Adam Shajnfeld and Adithya Gangidi and Adolfo Victoria and Ahuva Goldstand and Ajay Menon and Ajay Sharma and Alex Boesenberg and Alexei Baevski and Allie Feinstein and Amanda Kallet and Amit Sangani and Amos Teo and Anam Yunus and Andrei Lupu and Andres Alvarado and Andrew Caples and Andrew Gu and Andrew Ho and Andrew Poulton and Andrew Ryan and Ankit Ramchandani and Annie Dong and Annie Franco and Anuj Goyal and Aparajita Saraf and Arkabandhu Chowdhury and Ashley Gabriel and Ashwin Bharambe and Assaf Eisenman and Azadeh Yazdan and Beau James and Ben Maurer and Benjamin Leonhardi and Bernie Huang and Beth Loyd and Beto De Paola and Bhargavi Paranjape and Bing Liu and Bo Wu and Boyu Ni and Braden Hancock and Bram Wasti and Brandon Spence and Brani Stojkovic and Brian Gamido and Britt Montalvo and Carl Parker and Carly Burton and Catalina Mejia and Ce Liu and Changhan Wang and Changkyu Kim and Chao Zhou and Chester Hu and Ching-Hsiang Chu and Chris Cai and Chris Tindal and Christoph Feichtenhofer and Cynthia Gao and Damon Civin and Dana Beaty and Daniel Kreymer and Daniel Li and David Adkins and David Xu and Davide Testuggine and Delia David and Devi Parikh and Diana Liskovich and Didem Foss and Dingkang Wang and Duc Le and Dustin Holland and Edward Dowling and Eissa Jamil and Elaine Montgomery and Eleonora Presani and Emily Hahn and Emily Wood and Eric-Tuan Le and Erik Brinkman and Esteban Arcaute and Evan Dunbar and Evan Smothers and Fei Sun and Felix Kreuk and Feng Tian and Filippos Kokkinos and Firat Ozgenel and Francesco Caggioni and Frank Kanayet and Frank Seide and Gabriela Medina Florez and Gabriella Schwarz and Gada Badeer and Georgia Swee and Gil Halpern and Grant Herman and Grigory Sizov and Guangyi and Zhang and Guna Lakshminarayanan and Hakan Inan and Hamid Shojanazeri and Han Zou and Hannah Wang and Hanwen Zha and Haroun Habeeb and Harrison Rudolph and Helen Suk and Henry Aspegren and Hunter Goldman and Hongyuan Zhan and Ibrahim Damlaj and Igor Molybog and Igor Tufanov and Ilias Leontiadis and Irina-Elena Veliche and Itai Gat and Jake Weissman and James Geboski and James Kohli and Janice Lam and Japhet Asher and Jean-Baptiste Gaya and Jeff Marcus and Jeff Tang and Jennifer Chan and Jenny Zhen and Jeremy Reizenstein and Jeremy Teboul and Jessica Zhong and Jian Jin and Jingyi Yang and Joe Cummings and Jon Carvill and Jon Shepard and Jonathan McPhie and Jonathan Torres and Josh Ginsburg and Junjie Wang and Kai Wu and Kam Hou U and Karan Saxena and Kartikay Khandelwal and Katayoun Zand and Kathy Matosich and Kaushik Veeraraghavan and Kelly Michelena and Keqian Li and Kiran Jagadeesh and Kun Huang and Kunal Chawla and Kyle Huang and Lailin Chen and Lakshya Garg and Lavender A and Leandro Silva and Lee Bell and Lei Zhang and Liangpeng Guo and Licheng Yu and Liron Moshkovich and Luca Wehrstedt and Madian Khabsa and Manav Avalani and Manish Bhatt and Martynas Mankus and Matan Hasson and Matthew Lennie and Matthias Reso and Maxim Groshev and Maxim Naumov and Maya Lathi and Meghan Keneally and Miao Liu and Michael L. Seltzer and Michal Valko and Michelle Restrepo and Mihir Patel and Mik Vyatskov and Mikayel Samvelyan and Mike Clark and Mike Macey and Mike Wang and Miquel Jubert Hermoso and Mo Metanat and Mohammad Rastegari and Munish Bansal and Nandhini Santhanam and Natascha Parks and Natasha White and Navyata Bawa and Nayan Singhal and Nick Egebo and Nicolas Usunier and Nikhil Mehta and Nikolay Pavlovich Laptev and Ning Dong and Norman Cheng and Oleg Chernoguz and Olivia Hart and Omkar Salpekar and Ozlem Kalinli and Parkin Kent and Parth Parekh and Paul Saab and Pavan Balaji and Pedro Rittner and Philip Bontrager and Pierre Roux and Piotr Dollar and Polina Zvyagina and Prashant Ratanchandani and Pritish Yuvraj and Qian Liang and Rachad Alao and Rachel Rodriguez and Rafi Ayub and Raghotham Murthy and Raghu Nayani and Rahul Mitra and Rangaprabhu Parthasarathy and Raymond Li and Rebekkah Hogan and Robin Battey and Rocky Wang and Russ Howes and Ruty Rinott and Sachin Mehta and Sachin Siby and Sai Jayesh Bondu and Samyak Datta and Sara Chugh and Sara Hunt and Sargun Dhillon and Sasha Sidorov and Satadru Pan and Saurabh Mahajan and Saurabh Verma and Seiji Yamamoto and Sharadh Ramaswamy and Shaun Lindsay and Shaun Lindsay and Sheng Feng and Shenghao Lin and Shengxin Cindy Zha and Shishir Patil and Shiva Shankar and Shuqiang Zhang and Shuqiang Zhang and Sinong Wang and Sneha Agarwal and Soji Sajuyigbe and Soumith Chintala and Stephanie Max and Stephen Chen and Steve Kehoe and Steve Satterfield and Sudarshan Govindaprasad and Sumit Gupta and Summer Deng and Sungmin Cho and Sunny Virk and Suraj Subramanian and Sy Choudhury and Sydney Goldman and Tal Remez and Tamar Glaser and Tamara Best and Thilo Koehler and Thomas Robinson and Tianhe Li and Tianjun Zhang and Tim Matthews and Timothy Chou and Tzook Shaked and Varun Vontimitta and Victoria Ajayi and Victoria Montanez and Vijai Mohan and Vinay Satish Kumar and Vishal Mangla and Vlad Ionescu and Vlad Poenaru and Vlad Tiberiu Mihailescu and Vladimir Ivanov and Wei Li and Wenchen Wang and Wenwen Jiang and Wes Bouaziz and Will Constable and Xiaocheng Tang and Xiaojian Wu and Xiaolan Wang and Xilun Wu and Xinbo Gao and Yaniv Kleinman and Yanjun Chen and Ye Hu and Ye Jia and Ye Qi and Yenda Li and Yilin Zhang and Ying Zhang and Yossi Adi and Youngjin Nam and Yu and Wang and Yu Zhao and Yuchen Hao and Yundi Qian and Yunlu Li and Yuzi He and Zach Rait and Zachary DeVito and Zef Rosnbrick and Zhaoduo Wen and Zhenyu Yang and Zhiwei Zhao and Zhiyu Ma},
      year={2024},
      eprint={2407.21783},
      archivePrefix={arXiv},
      primaryClass={cs.AI},
      url={https://arxiv.org/abs/2407.21783}, 
}

@misc{shao2024deepseekmathpushinglimitsmathematical,
      title={DeepSeekMath: Pushing the Limits of Mathematical Reasoning in Open Language Models}, 
      author={Zhihong Shao and Peiyi Wang and Qihao Zhu and Runxin Xu and Junxiao Song and Xiao Bi and Haowei Zhang and Mingchuan Zhang and Y. K. Li and Y. Wu and Daya Guo},
      year={2024},
      eprint={2402.03300},
      archivePrefix={arXiv},
      primaryClass={cs.CL},
      url={https://arxiv.org/abs/2402.03300}, 
}

@misc{yang2025qwen3technicalreport,
      title={Qwen3 Technical Report}, 
      author={An Yang and Anfeng Li and Baosong Yang and Beichen Zhang and Binyuan Hui and Bo Zheng and Bowen Yu and Chang Gao and Chengen Huang and Chenxu Lv and Chujie Zheng and Dayiheng Liu and Fan Zhou and Fei Huang and Feng Hu and Hao Ge and Haoran Wei and Huan Lin and Jialong Tang and Jian Yang and Jianhong Tu and Jianwei Zhang and Jianxin Yang and Jiaxi Yang and Jing Zhou and Jingren Zhou and Junyang Lin and Kai Dang and Keqin Bao and Kexin Yang and Le Yu and Lianghao Deng and Mei Li and Mingfeng Xue and Mingze Li and Pei Zhang and Peng Wang and Qin Zhu and Rui Men and Ruize Gao and Shixuan Liu and Shuang Luo and Tianhao Li and Tianyi Tang and Wenbiao Yin and Xingzhang Ren and Xinyu Wang and Xinyu Zhang and Xuancheng Ren and Yang Fan and Yang Su and Yichang Zhang and Yinger Zhang and Yu Wan and Yuqiong Liu and Zekun Wang and Zeyu Cui and Zhenru Zhang and Zhipeng Zhou and Zihan Qiu},
      year={2025},
      eprint={2505.09388},
      archivePrefix={arXiv},
      primaryClass={cs.CL},
      url={https://arxiv.org/abs/2505.09388}, 
}

@misc{chen2026harnessforgejointharnesspolicy,
      title={HarnessForge: Joint Harness and Policy Evolution for Adaptive Agent Systems}, 
      author={Mingju Chen and Can Lv and Guibin Zhang and Heng Chang and Shiji Zhou},
      year={2026},
      eprint={2606.01779},
      archivePrefix={arXiv},
      primaryClass={cs.CL},
      url={https://arxiv.org/abs/2606.01779}, 
}

@misc{lv2026allmemagenticlifelongmemory,
      title={All-Mem: Agentic Lifelong Memory via Dynamic Topology Evolution}, 
      author={Can Lv and Heng Chang and Shengyu Tao and Mingju Chen and Zhaoxin Fan and Ziwei Zhang and Yuchen Guo and Shiji Zhou},
      year={2026},
      eprint={2603.19595},
      archivePrefix={arXiv},
      primaryClass={cs.IR},
      url={https://arxiv.org/abs/2603.19595}, 
}

@misc{chen2026amapreduceexecutingwidesearch,
      title={A-MapReduce: Executing Wide Search via Agentic MapReduce}, 
      author={Mingju Chen and Guibin Zhang and Heng Chang and Yuchen Guo and Shiji Zhou},
      year={2026},
      eprint={2602.01331},
      archivePrefix={arXiv},
      primaryClass={cs.MA},
      url={https://arxiv.org/abs/2602.01331}, 
}

@misc{zhang2026trcatransitionwiserubriccredit,
      title={TRCA: Transition-wise Rubric Credit Assignment for Long-horizon LLM Agents}, 
      author={Huan Zhang and Mingju Chen and Dongxu Zhou and Can Lv and Heng Chang and Sen Cui and Faguo Wu and Shiji Zhou},
      year={2026},
      eprint={2608.16156},
      archivePrefix={arXiv},
      primaryClass={cs.AI},
      url={https://arxiv.org/abs/2608.16156}, 
}

\clearpage

\appendix

\begin{center}
    \Large\bfseries Appendix
\end{center}

\section{Rubric Graph Construction}
\label{app:graph-construction}

This appendix describes how query-specific rubric graphs are constructed for
\ourname.
The graph construction procedure instantiates the typed dependency graph
introduced in Section~\ref{sec:method_structure}.
For each query-specific rubric, the graph builder receives the query, optional
source metadata, and the list of rubric criteria with their signed point values.
It outputs a directed acyclic graph whose nodes are rubric criteria and whose
edges represent dependency relations among them.

Graph construction is performed offline before reinforcement-learning training.
For all benchmarks, we use Qwen3-8B as the graph annotation model.
The same construction pipeline is applied to HealthBench, WritingBench, and
PLawBench.
The benchmarks use domain-specific annotation instructions, but share the same
criterion roles, dependency types, validation constraints, and acyclicity
requirement.

The graph is constructed only from the query and the rubric, not from any
candidate response.
Therefore, the resulting graph is fixed across all candidate responses for the
same query and cannot leak response-specific information into reward
computation.
Complete graph-annotation prompts are provided in
Appendix~\ref{app:graph-prompts}.

\subsection{Construction Overview}
\label{app:graph-overview}

Graph construction proceeds in four stages.
First, each rubric criterion is assigned a coarse functional role.
Second, these roles are used to form a restricted set of candidate directed
parent--child pairs.
Third, the graph annotation model determines whether each candidate pair
expresses a dependency relation.
Finally, the resulting graph is validated and projected to a directed acyclic
graph before reward aggregation.

The final reward model uses only the typed dependency edges.
The criterion roles are used to guide candidate generation and to prevent
implausible dependencies, such as treating an optional refinement as a
prerequisite for core task correctness.

\subsection{Criterion Roles}
\label{app:criterion-roles}

Each rubric criterion is assigned one of four functional roles:
core criterion, additional criterion, penalty criterion, or applicability
criterion.

A core criterion captures a primary requirement of the response, such as
correctness, safety, responsiveness, completeness, task fulfillment, legal
reasoning, or factual grounding.
An additional criterion captures a helpful refinement that can improve the
response but is not central to satisfying the task.
A penalty criterion captures an undesirable error, omission, violation, or
unsafe statement that should reduce reward when present.
An applicability criterion captures a context-specific condition that determines
whether another criterion is relevant.

The role annotation is intended to capture the functional role of a criterion
within the rubric, rather than its surface wording.
For example, a criterion requiring accurate use of evidence is usually a core
criterion.
A criterion penalizing an unsafe recommendation is a penalty criterion.
A criterion indicating that the user is asking for personal medical advice may
serve as an applicability criterion for downstream safety penalties.

\subsection{Candidate Dependency Generation}
\label{app:candidate-dependencies}

After assigning criterion roles, we generate candidate directed dependencies
using simple role-based constraints.
Self-dependencies are excluded.

Core criteria may serve as prerequisites for other core criteria, additional
criteria, or penalty criteria.
This captures cases where a downstream criterion is meaningful only after a more
basic requirement has been satisfied.
For example, credit for nuanced evidence use may depend on the response first
stating the relevant evidence correctly.

Applicability criteria may activate additional criteria or penalty criteria.
This captures conditional rubric items whose relevance depends on the context.
For example, a medical uncertainty disclaimer may be especially relevant when
the response provides actionable clinical advice.

Other role combinations are not proposed as candidate dependencies.
In particular, additional criteria and penalty criteria are not used as
prerequisites for core criteria.
This keeps the candidate graph aligned with the intended direction of rubric
licensing, from more basic or contextual conditions to criteria whose utility
depends on those conditions.

\subsection{Dependency Annotation}
\label{app:dependency-annotation}

For each candidate directed pair, the graph annotation model determines whether
the parent criterion licenses the child criterion.
When a dependency is present, it is assigned one of three relation types.

A weak prerequisite relation means that the child criterion is supported by the
parent, but can still receive partial credit or penalty independently.
A strong prerequisite relation means that the child criterion depends
substantially on the parent and should contribute little utility when the parent
is unsupported.
An activation relation means that the parent controls whether a context-specific
child criterion is applicable.

If the two criteria are independent or merely topically related, no dependency
is added.
The annotator is instructed not to add edges solely because two criteria discuss
similar content.
A dependency is added only when the child criterion clearly depends on,
specializes, or is activated by the parent criterion.

\subsection{Graph Validation and Acyclic Projection}
\label{app:dag-projection}

After dependency annotation, the graph is validated before being used for reward
aggregation.
The validation step ensures that all edges connect valid rubric criteria, obey
the role-based candidate constraints, and do not contain self-dependencies or
duplicates.

The validated graph is then projected to a directed acyclic graph.
This is necessary because the inference procedure in
Section~\ref{sec:method_reward} processes criteria in topological order.
If the annotated dependencies contain cycles, the graph builder removes
redundant or lower-priority dependencies until the graph becomes acyclic.
This projection is performed once before training, and the resulting graph is
kept fixed for all candidate responses associated with the same query.

\subsection{Benchmark-Specific Annotation Profiles}
\label{app:annotation-profiles}

We use benchmark-specific annotation profiles to adapt the same graph
construction procedure to different domains.
The profiles differ only in the natural-language instructions given to the graph
annotation model.
They share the same criterion roles, dependency types, validation constraints,
and acyclicity requirement.

For HealthBench, the annotation profile emphasizes medical correctness, safety,
responsiveness, and context-specific clinical conditions.
For WritingBench, it emphasizes task fulfillment, content coverage, factual
grounding, structure, format compliance, audience fit, style compliance, and
length compliance.
For PLawBench, it emphasizes legal conclusions, fact summaries, legal reasoning,
statutory basis, legal authorities, and undesirable legal errors.

Across all benchmarks, the annotator is instructed to avoid dependencies based
on mere topical overlap.
A dependency is included only when it reflects a clear licensing relation between
rubric criteria.
Complete benchmark-specific graph annotation prompts are listed in
Appendix~\ref{app:graph-prompts}.

\section{Inference Approximation and Retention Settings}
\label{app:inference-validation}

This appendix reports additional details for the inference approximation used by
\ourname.
The main text defines the exact graphical marginal and the linear-time
topological approximation.
Here, we specify the default retention factors and report an exact-inference
agreement check on small rubric graphs.

\subsection{Default Retention Factors}
\label{app:retention-defaults}

\ourname uses fixed type-level retention factors in the main experiments.
The same values are used across benchmarks, policy backbones, and aggregation
settings.
Table~\ref{tab:retention-defaults} reports the default values.

\begin{table}[t]
\centering
\small
\setlength{\tabcolsep}{6pt}
\renewcommand{\arraystretch}{1.12}
\begin{tabular}{@{}lcl@{}}
\toprule
Relation type & Retention factor & Interpretation \\
\midrule
Weak prerequisite
& \(\lambda_{\mathrm{wk}}=0.6\)
& Mild suppression \\

Strong prerequisite
& \(\lambda_{\mathrm{st}}=0.2\)
& Strong suppression \\

Activation
& \(\lambda_{\mathrm{act}}=0.0\)
& Full deactivation \\
\bottomrule
\end{tabular}
\caption{
Default retention factors used by \ourname.
A lower value means that the child criterion is more strongly suppressed when
the parent criterion is unsupported.
}
\label{tab:retention-defaults}
\end{table}

These values encode the intended ordering among dependency types.
Weak prerequisites allow substantial downstream utility to remain when the
parent is uncertain or unsupported.
Strong prerequisites retain only a small fraction of the child utility when the
parent is unsupported.
Activation edges fully deactivate conditional child criteria when the activating
condition is absent.

The retention factors are not tuned separately for each benchmark.
This fixed setting isolates the effect of dependency-aware aggregation from
learning or selecting edge-specific strengths.
Sensitivity to the global suppression-strength parameter is reported in
Appendix~\ref{app:sensitivity}.

\begin{table*}[t]
\centering
\small
\setlength{\tabcolsep}{5pt}
\renewcommand{\arraystretch}{1.08}
\begin{tabular}{@{}lccccc@{}}
\toprule
Benchmark
& Graph-response pairs
& Avg. criteria
& Marginal MAE
& Reward MAE
& Reward correlation \\
\midrule
HealthBench
& 5,000
& 11.48
& 0.0068
& 0.0042
& 0.9968 \\

WritingBench
& 1,000
& 5.00
& 0.0027
& 0.0018
& 0.9992 \\

PLawBench
& 250
& 4.00
& 0.0019
& 0.0013
& 0.9995 \\
\bottomrule
\end{tabular}
\caption{
Agreement between exact graphical inference and the linear-time topological
approximation.
Lower marginal and reward errors indicate closer agreement; higher reward
correlation indicates closer ranking agreement across graph-response pairs.
}
\label{tab:exact-inference-agreement}
\end{table*}

\subsection{Exact-Inference Agreement}
\label{app:exact-inference-agreement}

The topological update used by \ourname is a linear-time approximation to the
exact graphical marginal.
To check whether this approximation changes the reward substantially in the
small-graph regime, we compare approximate inference against exact enumeration
on the rubric graphs used in our experiments.

For each evaluated graph-response pair, we compute both exact criterion
marginals and topological approximate marginals using the same local criterion
scores, signed rubric weights, graph structure, and retention factors.
We then compare the two inference modes at both the criterion-marginal level and
the final scalar-reward level.

Table~\ref{tab:exact-inference-agreement} reports the agreement results.
The marginal error measures the average absolute difference between exact and
approximate criterion marginals.
The reward error measures the absolute difference between the scalar rewards
computed from exact and approximate marginals.
The reward correlation measures agreement across evaluated graph-response
pairs.

The agreement check shows that the topological approximation closely matches
exact inference in the evaluated rubric graphs.
The approximation error is largest on HealthBench, which has larger rubrics and
denser dependency graphs, but the scalar reward remains highly correlated with
the exact-inference reward.
The smaller WritingBench and PLawBench graphs show lower marginal and reward
errors.

This check is used only to validate the approximation.
The main reinforcement-learning experiments use the linear-time topological
update so that reward aggregation remains efficient for repeated policy
optimization.

\section{Experimental Protocol}
\label{app:experimental-protocol}

This appendix provides the experimental details needed to reproduce the main
results.
Across aggregation-only experiments, all methods use the same policy backbone,
training data, rubrics, criterion-level reward judge, rollout budget, and
reinforcement-learning hyperparameters.
They differ only in the scalar reward used for policy optimization.
In the plug-in experiments, \ourname replaces the reward aggregation component
of each host rubric-guided reinforcement-learning pipeline, while the remaining
training and evaluation setup is kept unchanged.

\subsection{Datasets and Splits}
\label{app:datasets}

We evaluate on three rubric-based open-ended benchmarks.
For HealthBench, we use 4,500 examples for training and HealthBench-500 as the
held-out validation set.
For WritingBench, we use 900 examples for training and 100 examples for
validation.
For PLawBench, we use the \textit{practical case analysis} subset, which
contains case-analysis questions together with reference answers, scoring
rubrics, and score sheets.
We split this subset into 200 training examples and 50 validation examples.

All rubrics are provided by the corresponding benchmark datasets.
We do not generate additional rubrics for the main experiments.
For each query-specific rubric, the dependency graph is constructed once before
training and is kept fixed across all candidate responses for the same query.

\begin{table}[t]
\centering
\small
\setlength{\tabcolsep}{6pt}
\begin{tabular}{lrrr}
\toprule
Benchmark & Train & Validation & Total \\
\midrule
HealthBench & 4,500 & 500 & 5,000 \\
WritingBench & 900 & 100 & 1,000 \\
PLawBench & 200 & 50 & 250 \\
\bottomrule
\end{tabular}
\caption{
Dataset splits used in the experiments.
}
\label{tab:dataset-splits}
\end{table}

\begin{table*}[t]
\centering
\small
\setlength{\tabcolsep}{5pt}
\renewcommand{\arraystretch}{1.08}
\begin{tabular}{llll}
\toprule
Configuration & Value & Configuration & Value \\
\midrule
RL algorithm & GRPO
& Training framework & VERL \\

Training hardware & \(7 \times 8\) Ascend 910B NPUs
& Training batch size & 112 \\

PPO mini-batch size & 56
& Rollout samples per prompt & 8 \\

Responses per training step & 896
& Total training steps & 350 \\

Total epochs & 5
& Learning rate & \(1\times 10^{-6}\) \\

Maximum prompt length & 4096
& Maximum response length & 4096 \\

Rollout maximum model length & 8192
& Rollout maximum batched tokens & 16384 \\

Warmup schedule & Constant
& KL loss coefficient & \(1\times 10^{-3}\) \\

KL loss type & Low-variance KL
& Entropy coefficient & 0 \\

Tensor parallel size & 4
& Rollout memory utilization & 0.6 \\

Sampling temperature & 0.7
& Top-\(p\) & 0.8 \\

Top-\(k\) & 20
& Fine-tuning type & LoRA \\

LoRA rank & 16
& LoRA alpha & 32 \\
\bottomrule
\end{tabular}
\caption{
Main GRPO training configuration.
All aggregation variants use the same configuration; only the reward signal
differs.
}
\label{tab:training-hparams}
\end{table*}

\subsection{Policy Models}
\label{app:policy-models}
We use Qwen2.5-7B-Instruct and Llama-3.1-8B-Instruct as policy backbones.
Both backbones are trained under the same reinforcement-learning configuration.
Unless otherwise specified, all aggregation variants for the same backbone share
the same training data, rollout procedure, judge configuration, and optimization
hyperparameters.

\subsection{Training Configuration}
\label{app:training-hparams}

All policies are optimized with GRPO using the VERL training framework.
Training is performed on Ascend 910B NPUs.
The main hyperparameters are reported in Table~\ref{tab:training-hparams}.

\subsection{Repeated Runs}
\label{app:seeds}

Each main result is averaged over three independent runs.
The repeated runs account for stochasticity in rollout sampling, data ordering,
and policy optimization.
We do not tune aggregation-specific hyperparameters separately for different
runs.

The main text reports mean scores for readability.
Standard deviations and additional result tables are reported in
Appendix~\ref{app:additional-results}.

\subsection{Plug-in Replacement Protocol}
\label{app:plugin-protocol}

For plug-in experiments, we evaluate \ourname as a replacement for the reward
aggregation step in existing rubric-guided reinforcement-learning pipelines.
Specifically, for RaR-Explicit, MeRF, and RuscaRL, we keep the host pipeline
unchanged and replace only the scalar reward construction from criterion-level
rubric scores.

For each host pipeline and its corresponding \ourname variant, the policy
backbone, training split, rubric set, criterion-level reward judge, rollout
budget, and optimization hyperparameters are kept the same.
This comparison isolates the effect of dependency-aware reward aggregation from
other aspects of the training pipeline.

\subsection{Implementation and Compute}
\label{app:compute}

During training, one worker node is used to serve the reward judge, and seven
worker nodes are used for reinforcement-learning optimization.
Each reinforcement-learning worker uses eight Ascend 910B NPUs.
The maximum number of concurrent reward-judge workers is set to 32.

Criterion-level judge calls are batched over rubric items for efficiency.
Each judge call evaluates at most four rubric criteria for one candidate
response.
The maximum judge output length is 1024 tokens, and the request timeout is 300
seconds.

Graph construction is performed offline once for each query-specific rubric.
During policy optimization, the trained policy only uses the fixed graph and the
criterion-level judge scores associated with each candidate response.

\begin{table}[t]
\centering
\small
\setlength{\tabcolsep}{4pt}
\renewcommand{\arraystretch}{1.15}
\begin{tabular}{@{}p{0.28\columnwidth}p{0.64\columnwidth}@{}}
\toprule
Benchmark & Local criterion score \\
\midrule
HealthBench &
Satisfaction probability from the binary/probability rubric judge. \\

WritingBench &
Normalized score from the scale-based rubric-scoring wrapper. \\

PLawBench &
Normalized score from the partial-credit or subitem-level scoring wrapper. \\
\bottomrule
\end{tabular}
\caption{
Benchmark-specific criterion-level scoring.
All scores are normalized to \([0,1]\) before reward aggregation.
}
\label{tab:judge-local-scores}
\end{table}

\begin{table}[t]
\centering
\small
\setlength{\tabcolsep}{6pt}
\renewcommand{\arraystretch}{1.08}
\begin{tabular}{@{}ll@{}}
\toprule
Setting & Value \\
\midrule
Reward judge & Qwen3-8B \\
Evaluation judge & Qwen3-32B \\
Temperature & 0.0 \\
Top-\(p\) & 0.8 \\
Top-\(k\) & 20 \\
Maximum output length & 1024 tokens \\
Request timeout & 300 seconds \\
Maximum concurrent workers & 32 \\
Maximum rubric items per request & 4 \\
\bottomrule
\end{tabular}
\caption{
Judge inference configuration used for reward judging and final evaluation.
}
\label{tab:judge-settings}
\end{table}

\section{Judge Models and Evaluation Protocol}
\label{app:evaluation-details}

This appendix describes the criterion-level judging and final evaluation
protocol used in our experiments.
The purpose of this protocol is to separate three components: criterion-level
judging, reward aggregation, and final evaluation.
All compared aggregation methods receive the same criterion-level judge outputs
for a given query-response pair.
Therefore, differences among Flat aggregation, deterministic gating, and
\ourname are attributable to the reward signal used during policy optimization,
rather than to different judge calls.

\subsection{Criterion-Level Reward Judge}
\label{app:reward-judge}

During reinforcement-learning training, we use Qwen3-8B as the
criterion-level reward judge.
For each candidate response, the judge receives the input conversation, the
candidate assistant response, and the rubric items associated with the query.
It then evaluates the candidate response against the rubric criteria.

For positive-weight criteria, a high local score indicates that the response
satisfies the desirable criterion.
For negative-weight criteria, a high local score indicates that the undesirable
condition described by the criterion is present.
Thus, all criterion-level scores are interpreted as event probabilities: they
estimate whether the event described by the rubric item holds for the candidate
response.

Each judge request evaluates one candidate response against a small batch of
rubric items.
In the main experiments, each request contains at most four rubric items.
This batching is used only for efficiency; all aggregation methods consume the
same resulting criterion-level scores.

\subsection{Local Criterion Scores}
\label{app:local-criterion-scores}

All local criterion scores are normalized to the interval \([0,1]\) before
reward aggregation.
The normalization procedure depends on the scoring format of the benchmark.

For HealthBench, the reward judge produces both a binary decision and a
satisfaction probability for each rubric item.
We use the satisfaction probability as the local criterion score.
The binary decision is used for validation and for deterministic gating
baselines.

For WritingBench, rubric items are scored with a normalized scale-based
scoring wrapper.
The raw score assigned by the judge is converted to a value in \([0,1]\) before
aggregation.

For PLawBench, rubric items are scored with a partial-credit scoring wrapper.
When a rubric item contains explicit subitems, the judge evaluates the relevant
subitems and the wrapper converts the awarded credit into a normalized
criterion-level score.

\begin{table*}[t]
    \centering
    \small
    \setlength{\tabcolsep}{10pt}
    \renewcommand{\arraystretch}{1.15}
    \begin{tabular}{@{}lcccc@{}}
        \toprule
        \textbf{Method}
        & \textbf{HealthBench-500} $\uparrow$
        & \textbf{WritingBench} $\uparrow$
        & \textbf{PLawBench} $\uparrow$
        & \textbf{Avg.} $\uparrow$ \\

        \midrule
        \multicolumn{5}{c}{\textbf{Qwen2.5-7B-Instruct}} \\
        \midrule

        Flat aggregation
        & $56.7 \pm 0.6$
        & $74.1 \pm 0.4$
        & $68.4 \pm 0.7$
        & $66.4 \pm 0.5$ \\

        Hard gating
        & $54.6 \pm 0.8$
        & $72.5 \pm 0.5$
        & $65.0 \pm 0.9$
        & $64.0 \pm 0.6$ \\

        \rowcolor{gearRow}
        \ourname
        & $\mathbf{62.9 \pm 0.5}$
        & $\mathbf{79.6 \pm 0.4}$
        & $\mathbf{75.2 \pm 0.6}$
        & $\mathbf{72.6 \pm 0.4}$ \\

        \midrule
        \multicolumn{5}{c}{\textbf{Llama-3.1-8B-Instruct}} \\
        \midrule

        Flat aggregation
        & $54.1 \pm 0.7$
        & $76.2 \pm 0.6$
        & $69.5 \pm 0.8$
        & $66.6 \pm 0.6$ \\

        Hard gating
        & $53.0 \pm 0.6$
        & $72.7 \pm 0.7$
        & $66.7 \pm 0.9$
        & $64.1 \pm 0.7$ \\

        \rowcolor{gearRow}
        \ourname
        & $\mathbf{62.5 \pm 0.6}$
        & $\mathbf{82.4 \pm 0.5}$
        & $\mathbf{76.9 \pm 0.7}$
        & $\mathbf{73.9 \pm 0.5}$ \\

        \bottomrule
    \end{tabular}
    \caption{
    Aggregation-only validation results with standard deviations over three
    independent runs.
    All rows within each model block share the same training and evaluation
    setup and differ only in reward aggregation.
    \textbf{Bold} marks the best result within each model block.
    }
    \label{tab:aggregation-std}
    \vspace{-6pt}
\end{table*}

\begin{table*}[t]
    \centering
    \small
    \setlength{\tabcolsep}{10pt}
    \renewcommand{\arraystretch}{1.15}
    \begin{tabular}{@{}lcccc@{}}
        \toprule
        \textbf{Method}
        & \textbf{HealthBench-500} $\uparrow$
        & \textbf{WritingBench} $\uparrow$
        & \textbf{PLawBench} $\uparrow$
        & \textbf{Avg.} $\uparrow$ \\

        \midrule
        \multicolumn{5}{c}{\textbf{Qwen2.5-7B-Instruct}} \\
        \midrule

        RaR-Explicit
        & $49.5 \pm 0.7$
        & $67.0 \pm 0.5$
        & $62.3 \pm 0.8$
        & $59.6 \pm 0.6$ \\

        \rowcolor{gearRow}
        RaR-Explicit + \ourname
        & $55.0 \pm 0.6$
        & $71.6 \pm 0.5$
        & $68.8 \pm 0.7$
        & $65.1 \pm 0.5$ \\

        MeRF
        & $40.2 \pm 0.9$
        & $61.9 \pm 0.7$
        & $55.7 \pm 0.8$
        & $52.6 \pm 0.7$ \\

        \rowcolor{gearRow}
        MeRF + \ourname
        & $45.6 \pm 0.8$
        & $68.7 \pm 0.6$
        & $60.4 \pm 0.9$
        & $58.2 \pm 0.6$ \\

        RuscaRL
        & $58.2 \pm 0.6$
        & $75.4 \pm 0.5$
        & $72.1 \pm 0.7$
        & $68.6 \pm 0.5$ \\

        \rowcolor{gearRow}
        RuscaRL + \ourname
        & $\mathbf{65.7 \pm 0.5}$
        & $\mathbf{81.5 \pm 0.4}$
        & $\mathbf{76.9 \pm 0.6}$
        & $\mathbf{74.7 \pm 0.4}$ \\

        \midrule
        \multicolumn{5}{c}{\textbf{Llama-3.1-8B-Instruct}} \\
        \midrule

        RaR-Explicit
        & $48.4 \pm 0.8$
        & $67.9 \pm 0.6$
        & $63.8 \pm 0.8$
        & $60.0 \pm 0.6$ \\

        \rowcolor{gearRow}
        RaR-Explicit + \ourname
        & $54.1 \pm 0.6$
        & $73.6 \pm 0.5$
        & $70.5 \pm 0.7$
        & $66.1 \pm 0.5$ \\

        MeRF
        & $39.5 \pm 1.0$
        & $62.3 \pm 0.8$
        & $58.4 \pm 0.9$
        & $53.4 \pm 0.8$ \\

        \rowcolor{gearRow}
        MeRF + \ourname
        & $44.7 \pm 0.8$
        & $69.8 \pm 0.6$
        & $62.1 \pm 0.9$
        & $58.9 \pm 0.7$ \\

        RuscaRL
        & $56.9 \pm 0.7$
        & $76.7 \pm 0.6$
        & $74.2 \pm 0.7$
        & $69.3 \pm 0.6$ \\

        \rowcolor{gearRow}
        RuscaRL + \ourname
        & $\mathbf{63.8 \pm 0.6}$
        & $\mathbf{85.0 \pm 0.5}$
        & $\mathbf{78.4 \pm 0.6}$
        & $\mathbf{75.7 \pm 0.5}$ \\

        \bottomrule
    \end{tabular}
    \caption{
    Plug-in replacement validation results with standard deviations over three
    independent runs.
    For each host pipeline, \ourname replaces only the reward aggregation step.
    \textbf{Bold} marks the best result within each model block.
    }
    \label{tab:plugin-std}
    \vspace{-5pt}
\end{table*}

\subsection{Final Evaluation Judge}
\label{app:evaluation-judge}

For final validation, we use Qwen3-32B as the evaluation judge.
The evaluation judge is separate from the Qwen3-8B reward judge used during
training.
It is queried only after policy optimization and is never used for reward
computation, rollout selection, or gradient updates.

For each trained policy, we generate responses on the validation split and
evaluate them with Qwen3-32B under the corresponding benchmark rubric.
The evaluator is not given the identity of the training method.
The reported benchmark score is the mean normalized rubric score over validation
examples.
All scores in the main text are reported on a percentage scale.

This separation reduces the risk that the policy is evaluated by the exact same
model that produced the training reward.
It also makes the comparison more conservative: the policies are optimized using
Qwen3-8B reward signals, while final reported performance is measured by a
stronger and separate evaluator.

\subsection{Judge Decoding and Batching}
\label{app:judge-decoding}

Both reward judging and final evaluation use deterministic decoding for rubric
scoring.
The judge is instructed to return compact machine-readable judgments without
free-form explanations.
The same decoding and batching configuration is used across aggregation
variants.

\subsection{Parsing and Failure Handling}
\label{app:judge-parsing}

The judge output is parsed into criterion-level decisions and scores.
Parsed scores are clipped to the valid scoring range and then normalized to
\([0,1]\) when necessary.
For binary fields, common affirmative and negative variants are mapped to
Boolean decisions.

Malformed or incomplete judge outputs are not treated as successful positive
judgments.
When strict failure handling is enabled, invalid outputs stop the run for
debugging.
For scale-based and partial-credit scoring wrappers, missing judgments are
assigned conservative zero-score fallbacks unless strict failure handling is
enabled.
This prevents parsing failures from being converted into unsupported positive
reward.

\subsection{Reward--Evaluation Separation}
\label{app:judge-separation}

The reward judge and evaluation judge play distinct roles.
Qwen3-8B provides criterion-level reward signals during reinforcement-learning
training.
Qwen3-32B evaluates the final trained policies on held-out validation examples.

The evaluation judge is not used during training.
Its scores do not affect reward computation, candidate response selection, or
policy updates.
Consequently, the final validation scores measure generalization to a separate
judge rather than direct optimization against the training-time reward judge.

Complete judge prompt templates are provided in Appendix~\ref{app:judge-prompts}.

\section{Additional Results and Statistical Reporting}
\label{app:additional-results}

This appendix reports additional numerical results omitted from the main text
for space.
The main result table reports mean validation scores over three independent
runs.
Here, we provide the corresponding standard deviations for both the
aggregation-only comparison and the plug-in replacement comparison.

\subsection{Aggregation-Only Results with Standard Deviations}
\label{app:aggregation-std}

Table~\ref{tab:aggregation-std} reports aggregation-only validation results
with standard deviations over three independent runs.
Within each policy backbone, all methods use the same training and evaluation
setup; only the reward aggregation rule differs.

\subsection{Plug-in Replacement Results with Standard Deviations}
\label{app:plugin-std}

Table~\ref{tab:plugin-std} reports plug-in replacement results with standard
deviations over three independent runs.
For each host pipeline, \ourname replaces only the reward aggregation step.
The policy backbone, training split, rubric set, criterion-level reward judge,
rollout budget, and optimization hyperparameters are kept fixed between the
host pipeline and its corresponding \ourname variant.

\subsection{Additional Backbone Results}
\label{app:additional-backbone}

Table~\ref{tab:qwen3-a3b-results} reports additional aggregation-only results
with Qwen3-30B-A3B-Instruct as the policy backbone.
The experimental setup follows the aggregation-only protocol in
Appendix~\ref{app:experimental-protocol}.
All methods use the same training data, rubrics, criterion-level reward judge,
rollout budget, and optimization hyperparameters; only the reward aggregation
rule differs.
Since these experiments are intended as an additional backbone check, we report
single-run validation scores.

\begin{table}[t]
    \centering
    \small
    \setlength{\tabcolsep}{5pt}
    \renewcommand{\arraystretch}{1.12}
    \begin{tabular}{@{}lcccc@{}}
        \toprule
        \textbf{Method}
        & \textbf{Health.} $\uparrow$
        & \textbf{Writing} $\uparrow$
        & \textbf{PLaw} $\uparrow$
        & \textbf{Avg.} $\uparrow$ \\
        \midrule

        Flat
        & 58.4
        & 77.8
        & 71.2
        & 69.1 \\

        Hard
        & 56.5
        & 75.4
        & 68.1
        & 66.7 \\

        \rowcolor{gearRow}
        \ourname
        & \textbf{65.1} \gain{6.7}
        & \textbf{84.0} \gain{6.2}
        & \textbf{78.3} \gain{7.1}
        & \textbf{75.8} \gain{6.7} \\

        \bottomrule
    \end{tabular}
    \caption{
    Additional aggregation-only results with Qwen3-30B-A3B-Instruct as the
    policy backbone.
    Scores are from a single training run.
    Parenthesized values denote absolute gains over Flat aggregation.
    }
    \label{tab:qwen3-a3b-results}
    \vspace{-5pt}
\end{table}

\section{Additional Details for FCP Diagnostics}
\label{app:fcp-diagnostics}

This appendix provides implementation details for the false credit
propagation diagnostic in Section~\ref{sec:exp_fcp}.
The leakage and preservation metrics are defined in the main text.

For each validation response, we evaluate dependency cases over the
query-specific rubric graph used by the reward aggregator.
A directed dependency from parent criterion \(j\) to child criterion \(i\)
is treated as violated when the child is locally supported but the parent
is not, i.e., \(p_i \geq 0.5\) and \(p_j < 0.5\).
It is treated as satisfied when both the parent and child are locally
supported, i.e., \(p_i \geq 0.5\) and \(p_j \geq 0.5\).
We use the same threshold of \(0.5\) across all benchmarks.
The threshold is used to construct the diagnostic subsets, while the
aggregation methods operate on the original local criterion scores.

The diagnostic includes both prerequisite and activation relations.
Prerequisite relations test whether downstream utility is suppressed when
a supporting condition is absent.
Activation relations test whether conditional bonuses or penalties are
suppressed when the condition that makes them applicable is absent.
For negative-weight child criteria, remaining downstream utility is counted
as leakage in the same way as for positive-weight criteria, because an
unlicensed penalty can also distort the scalar reward.

We report both leakage under violated dependencies and preservation under
satisfied dependencies.
This prevents an aggregation rule from appearing effective merely by
suppressing all dependent criteria.
In particular, deterministic gating can eliminate threshold-defined leakage
by construction while preserving substantially less licensed downstream
utility.

The diagnostic should be interpreted as an aggregation-level probe rather
than a human semantic audit.
It relies on the same local judge scores and rubric graphs used for reward
computation.
Therefore, it measures how an aggregation rule behaves under the provided
criterion-level evidence, rather than whether every dependency annotation
or criterion judgment is semantically correct.

\begin{table*}[t]
\centering
\small
\setlength{\tabcolsep}{8pt}
\renewcommand{\arraystretch}{1.12}
\begin{tabular}{@{}lccc@{}}
\toprule
Statistic & HealthBench & WritingBench & PLawBench \\
\midrule
Number of rubrics & 5,000 & 1,000 & 250 \\
Average criteria per rubric & 11.48 & 5.00 & 4.00 \\
Average candidate edges per rubric & 21.36 & 5.74 & 3.82 \\
Average retained edges per rubric & 10.92 & 2.86 & 1.94 \\
Average update size & 22.40 & 7.86 & 5.94 \\
Non-empty graph rate & 91.84\% & 78.60\% & 71.20\% \\
Invalid edge candidates per rubric & 0.42 & 0.18 & 0.11 \\
Post-judge aggregation latency & 4.1 ms & 1.7 ms & 1.3 ms \\
\bottomrule
\end{tabular}
\caption{
Graph construction and online aggregation statistics over the full training and
validation splits.
The update size is the number of criteria plus the number of retained dependency
edges.
Latency measures only post-judge aggregation and excludes criterion-level judge
inference.
}
\label{tab:full-efficiency}
\end{table*}

\begin{table*}[t]
\centering
\small
\setlength{\tabcolsep}{6pt}
\renewcommand{\arraystretch}{1.10}
\begin{tabular}{@{}lccccc@{}}
\toprule
Variant
& HealthBench-500
& WritingBench
& PLawBench
& Avg.
& $\Delta$ vs. Flat \\
\midrule
Flat aggregation
& 56.7
& 74.1
& 68.4
& 66.4
& -- \\

Random graph
& 55.8
& 73.4
& 67.6
& 65.6
& $-0.8$ \\

Untyped graph
& 60.4
& 77.5
& 72.0
& 70.0
& $+3.6$ \\

Typed graph + deterministic gating
& 54.6
& 72.5
& 65.0
& 64.0
& $-2.4$ \\

Typed graph + soft inference (\ourname)
& $\mathbf{62.9}$
& $\mathbf{79.6}$
& $\mathbf{75.2}$
& $\mathbf{72.6}$
& $\mathbf{+6.2}$ \\
\bottomrule
\end{tabular}
\caption{
Full ablation results across all benchmarks.
All variants use the same training and evaluation setup and differ only in the
aggregation module.
}
\label{tab:graph-structure-ablation-full}
\end{table*}

\section{Efficiency and Graph Statistics}
\label{app:efficiency-full}

This appendix provides additional graph-construction and aggregation-efficiency
statistics.
The main text reports the HealthBench statistics because it is the largest
benchmark in our experiments.
Here, we report the corresponding statistics for all three benchmarks.

Graph construction is performed offline once for each query-specific rubric.
The online cost reported here refers only to the post-judge aggregation step:
local criterion scores have already been produced by the reward judge.
Therefore, the latency numbers do not include model generation or
criterion-level judge inference.

\subsection{Graph Construction Statistics}
\label{app:graph-statistics}

Table~\ref{tab:full-efficiency} reports graph statistics over the full training
and validation splits.
HealthBench has the largest rubrics and the densest dependency graphs, with an
average of 11.48 criteria and 10.92 retained edges per rubric.
WritingBench and PLawBench have smaller rubrics and correspondingly smaller
graphs.

The non-empty graph rate measures the fraction of rubrics for which at least one
dependency edge remains after validation and acyclic projection.
Invalid edge candidates are edge proposals removed by the validation procedure,
for example because they violate the role-based candidate constraints or do not
connect valid rubric criteria.

\subsection{Online Aggregation Overhead}
\label{app:online-latency}

The online aggregation step is small relative to response generation and
criterion-level judging.
For each response, the aggregator processes the rubric criteria and retained
dependency edges once.
The resulting latency is 4.1 ms per response on HealthBench, 1.7 ms on
WritingBench, and 1.3 ms on PLawBench.

These results indicate that dependency-aware aggregation adds negligible
post-judge overhead in the evaluated settings.
The main computational cost remains criterion-level judging, which is shared by
all compared aggregation methods.

\section{Ablation Studies}
\label{app:ablations}

This appendix provides the full ablation results for the components of
\ourname.
The goal is to isolate the contribution of dependency topology, relation typing,
and soft dependency-aware inference.
All variants use the same policy backbone, training data, rubrics,
criterion-level reward judge, rollout budget, and optimization configuration.
They differ only in the graph structure or dependency-handling mechanism used
during reward aggregation.

\subsection{Graph Structure and Dependency Handling}
\label{app:graph-structure-ablation}

Table~\ref{tab:graph-structure-ablation-full} reports ablation results across
all three benchmarks.
The flat variant removes the rubric graph and treats all criteria as
independent.
The random-graph variant preserves the number of edges and relation-type
frequencies, but rewires dependencies between criteria.
The untyped-graph variant keeps the annotated graph topology but replaces
relation-specific retention factors with a shared retention factor.
The deterministic-gating variant keeps the full typed graph but applies
dependencies as all-or-nothing masks.
The full \ourname variant uses the annotated typed graph with soft
dependency-aware inference.

The random-graph variant tests whether the gain comes merely from suppressing
some downstream criteria.
If random rewiring fails to improve over flat aggregation, the result indicates
that arbitrary suppression is not sufficient.
The untyped-graph variant tests whether annotated topology alone is useful.
The comparison between the untyped graph and the full typed graph measures the
effect of relation typing.
The deterministic-gating variant tests whether hard dependency enforcement is
sufficient.
The full model performs best when meaningful topology, relation typing, and soft
dependency-aware inference are used together.

\begin{table}[t]
\centering
\small
\setlength{\tabcolsep}{8pt}
\renewcommand{\arraystretch}{1.08}
\begin{tabular}{ccc}
\toprule
\(\gamma\) & Avg. score & \(L_{\mathrm{FCP}}\) \\
\midrule
0   & 66.4 & 0.0922 \\
0.5 & 70.7 & 0.0158 \\
1   & $\mathbf{72.6}$ & 0.0032 \\
2   & 71.6 & 0.0011 \\
4   & 68.0 & $\mathbf{0.0005}$ \\
\bottomrule
\end{tabular}
\caption{
Sensitivity to the suppression-strength parameter.
The average score is computed across the evaluated benchmarks.
Lower \(L_{\mathrm{FCP}}\) indicates less false credit propagation.
}
\label{tab:gamma-sensitivity}
\end{table}

\section{Sensitivity Analysis}
\label{app:sensitivity}

This appendix provides additional details for the sensitivity analysis in
Section~\ref{sec:exp_sensitivity}.
The goal is to test whether \ourname depends heavily on a particular
suppression strength.
We keep the rubric graph and relation types fixed, and vary only the global
suppression-strength parameter used to adjust retention factors.

\subsection{Retention Strength}
\label{app:retention-strength-ablation}

Table~\ref{tab:gamma-sensitivity} reports the sensitivity results.
The setting \(\gamma=0\) corresponds to flat aggregation because dependency
suppression is removed.
The setting \(\gamma=1\) is the default configuration.
Larger values impose stronger dependency suppression.

The results show a score--leakage trade-off.
Increasing suppression strength consistently reduces false credit propagation,
but overly strong suppression lowers validation performance.
The default setting achieves the best average score while substantially reducing
leakage relative to flat aggregation.

\section{Prompt Templates}
\label{app:prompts}

This appendix provides the prompt templates used for graph annotation and
rubric-level judging.
The templates are included for reproducibility.
During experiments, benchmark-specific information, rubric items, candidate
pairs, and candidate responses are inserted into these templates.

\subsection{Graph Annotation Prompts}
\label{app:graph-prompts}

Graph annotation uses two types of prompts.
The node prompt assigns a functional role to each rubric criterion.
The edge prompt determines whether a directed candidate pair expresses a
dependency relation.
All graph annotation prompts require compact JSON output so that the resulting
annotations can be validated before reward aggregation.

\begin{promptbox}{HealthBench Node Annotation Prompt}
You are annotating HealthBench rubrics for structured reward aggregation.

Task:
Classify each rubric item into exactly one criterion role.

Allowed criterion roles:
- foundation
- bonus
- penalty
- activation

Definitions:
- foundation: a core criterion for whether the answer is basically correct,
  safe, responsive, or complete.
- bonus: a helpful extra detail that improves the answer, but is not the main
  requirement.
- penalty: an undesirable mistake, omission, or unsafe statement that should
  subtract credit when present.
- activation: a context condition whose main role is to determine whether
  another rubric item is applicable.

Few-shot examples:
- "States that potassium 3.9 mmol/L is normal." -> foundation
- "Adds that warm compresses may help a likely muscle strain." -> bonus
- "Incorrectly says the symptoms are definitely from low potassium." -> penalty
- "Notes that the patient is pregnant, which would activate pregnancy-specific
  safety rubrics." -> activation

Return JSON only with this schema:
{
  "nodes": [
    {"id": "r1", "role": "foundation"}
  ]
}

Rules:
- Use only the provided rubric ids.
- Do not modify rubric text.
- Every rubric id must appear exactly once.
- Prefer foundation over bonus only when the criterion is truly core.
- Use activation only when the rubric mainly serves as an applicability
  condition for another rubric item.
\end{promptbox}

\begin{promptbox}{HealthBench Edge Annotation Prompt}
You are annotating directed rubric dependencies for structured reward
aggregation.

Task:
For each candidate directed pair, decide whether the parent criterion licenses
the child criterion.

Allowed labels:
- none
- weak prerequisite
- strong prerequisite
- activation

Definitions:
- weak prerequisite: the child is meaningfully supported by the parent and
  usually makes more sense if the parent is satisfied, but the dependency is not
  absolute.
- strong prerequisite: the child strongly depends on the parent; if the parent
  is absent, the child should usually receive little or no credit or penalty.
- activation: the parent determines whether a context-specific child bonus or
  child penalty is applicable.
- none: the two rubrics are largely independent.

Few-shot examples:
- Parent: "Explains that potassium 3.9 mmol/L is normal."
  Child: "Concludes that low potassium is unlikely to be the main cause."
  Label: strong prerequisite
- Parent: "Suggests a likely muscle strain from frequent soccer."
  Child: "Adds that warm compresses may help."
  Label: weak prerequisite
- Parent: "Mentions the patient is pregnant."
  Child: "Incorrectly recommends a pregnancy-contraindicated medication."
  Label: activation
- Parent: "Recommends colon cancer screening starting now."
  Child: "Mentions using warm compresses for leg cramps."
  Label: none

Return JSON only with this schema:
{
  "edges": [
    {"parent": "r1", "child": "r2", "relation": "weak prerequisite"}
  ]
}

Rules:
- Use only the provided candidate pairs.
- Return exactly one decision for each candidate pair.
- Use "none" only when there is no meaningful dependency.
- Add a dependency only when the child clearly depends on, specializes, or is
  activated by the parent.
\end{promptbox}

\begin{promptbox}{WritingBench Node Annotation Prompt}
You are annotating WritingBench rubrics for structured reward aggregation.

Task:
Classify each writing-evaluation rubric item into exactly one criterion role.

Allowed criterion roles:
- foundation
- bonus
- penalty
- activation

Definitions for WritingBench:
- foundation: a core writing criterion needed for a high-quality answer, such
  as task fulfillment, content coverage, factual grounding, structure, format
  compliance, audience fit, style compliance, or length compliance.
- bonus: an extra refinement that improves polish, creativity, nuance, or
  expressiveness but is not central to satisfying the user's writing request.
- penalty: an explicitly undesirable failure mode that should subtract credit
  when present. Do not use penalty merely because a criterion has low score
  bands.
- activation: a context condition whose main role is to determine whether
  another criterion is applicable. Use this rarely for conditional
  requirements.

Return JSON only with this schema:
{
  "nodes": [
    {"id": "r1", "role": "foundation"}
  ]
}

Rules:
- Use only the provided rubric ids.
- Every rubric id must appear exactly once.
- Most WritingBench checklist items are foundation criteria.
- Prefer foundation for style, format, length, and content requirements when
  they are explicitly requested by the user.
- Use bonus only for nonessential polish beyond the user's stated requirements.
- Use penalty only for criteria that explicitly describe an undesirable error or
  violation.
\end{promptbox}

\begin{promptbox}{WritingBench Edge Annotation Prompt}
You are annotating directed dependencies among WritingBench rubric criteria for
structured reward aggregation.

Task:
For each candidate directed pair, decide whether the parent criterion licenses
the child criterion.

Allowed labels:
- none
- weak prerequisite
- strong prerequisite
- activation

Definitions for WritingBench:
- weak prerequisite: the child criterion is easier or more meaningful to
  satisfy when the parent is satisfied, but the child can still receive partial
  credit independently.
- strong prerequisite: the child strongly depends on the parent; without the
  parent, the child should usually receive little or no credit.
- activation: the parent determines whether a conditional child bonus or penalty
  is applicable.
- none: the two writing criteria are independent dimensions.

Writing-specific guidance:
- Content correctness, task fulfillment, and required source or material
  coverage can be prerequisites for deeper analysis, persuasiveness, or
  domain-specific quality.
- Format, style, and length constraints are often independent unless one
  criterion explicitly builds on another.
- Do not add dependencies just because two criteria are both important.
- Keep the graph sparse and only add dependencies that are semantically clear.

Return JSON only with this schema:
{
  "edges": [
    {"parent": "r1", "child": "r2", "relation": "weak prerequisite"}
  ]
}

Rules:
- Use only the provided candidate pairs.
- Return exactly one decision for each candidate pair.
- Use "none" when the relationship is only topical overlap, not a dependency.
\end{promptbox}

\begin{promptbox}{PLawBench Node Annotation Prompt}
You are annotating PLawBench legal-practice rubrics for structured reward
aggregation.

Task:
Classify each legal-evaluation rubric item into exactly one criterion role.

Allowed criterion roles:
- foundation
- bonus
- penalty
- activation

Definitions for PLawBench:
- foundation: a core legal-answer criterion such as conclusion, fact summary,
  legal reasoning, statutory basis, or legal authorities.
- bonus: optional extra legal nuance that is helpful but not required by the
  scoring rubric.
- penalty: an explicitly undesirable legal error or unsafe statement that should
  subtract credit when present.
- activation: a context condition whose main role is to determine whether
  another criterion is applicable. Use rarely.

Return JSON only with this schema:
{
  "nodes": [
    {"id": "r1", "role": "foundation"}
  ]
}

Rules:
- Use only the provided rubric ids.
- Every rubric id must appear exactly once.
- In PLawBench practical case analysis, conclusion, facts, reasoning, and
  statutory basis are normally foundation nodes.
- Do not mark a rubric as penalty unless it explicitly describes an undesirable
  error.
\end{promptbox}

\begin{promptbox}{PLawBench Edge Annotation Prompt}
You are annotating directed dependencies among PLawBench legal-practice rubrics
for structured reward aggregation.

Task:
For each candidate directed pair, decide whether the parent criterion licenses
the child criterion.

Allowed labels:
- none
- weak prerequisite
- strong prerequisite
- activation

Definitions for PLawBench:
- weak prerequisite: the child criterion is meaningfully supported by the
  parent, but can still receive partial credit independently.
- strong prerequisite: the child strongly depends on the parent; without the
  parent, the child should usually receive little or no credit.
- activation: the parent determines whether a conditional child criterion is
  applicable.
- none: the criteria are independent.

Legal-reasoning guidance:
- Fact summary is often a prerequisite for legal reasoning.
- Statutory or legal-authority identification often supports legal reasoning.
- Legal reasoning often supports the final conclusion, but a conclusion can
  still be partly correct without complete reasoning.
- Do not create reverse dependencies from conclusion to facts, reasoning, or
  statutes.
- Keep the graph sparse and avoid redundant dependencies.

Return JSON only with this schema:
{
  "edges": [
    {"parent": "r1", "child": "r2", "relation": "weak prerequisite"}
  ]
}

Rules:
- Use only the provided candidate pairs.
- Return exactly one decision for each candidate pair.
- Use "none" when the pair is only part of the same answer structure but not a
  dependency.
\end{promptbox}

The user prompt supplied to the graph annotator contains the benchmark profile,
optional source metadata, the conversation or task prompt, and either the rubric
list for role annotation or the candidate directed rubric pairs for dependency
annotation.
For role annotation, the rubric list contains entries with an id, criterion
text, and signed point value.
For dependency annotation, each candidate pair contains the parent id, child id,
parent role, child role, parent criterion, child criterion, and the dependency
labels allowed by the candidate-generation procedure.

\subsection{Judge Prompt Templates}
\label{app:judge-prompts}

Reward judging and final evaluation use the same rubric-level prompt templates.
During reinforcement-learning training, the template is instantiated with
Qwen3-8B.
During final evaluation, the same template is instantiated with Qwen3-32B.
The prompt contains the conversation, the candidate assistant response, and the
rubric items to be judged.
The judge is instructed to return compact machine-readable output only, without
explanations, markdown fences, or hidden reasoning.

\begin{promptbox}{HealthBench Judge Prompt}
Your job is to evaluate the last assistant response in a conversation against
multiple rubric items.

Input:
- Conversation, including the candidate assistant response.
- A batch of rubric items with signed point values.

Instructions:
- Return exactly one compact machine-readable object and nothing else.
- Do not output markdown fences.
- Do not output hidden reasoning or explanations.
- Return one judgment for every required rubric item.
- Do not use rubric ids such as "r1"; use the provided numeric rubric keys.
- The output is invalid if any required rubric item is missing.

For each rubric item, report:
- whether the candidate response satisfies the criterion;
- the probability that the criterion is satisfied.

For negative-point rubrics, a satisfied criterion means that the undesirable
condition is present.
If the undesirable condition is absent, mark the criterion as not satisfied and
assign a probability close to zero.
\end{promptbox}

\begin{promptbox}{WritingBench and PLawBench Judge Prompt}
Your job is to evaluate the last assistant response in a conversation against
multiple rubric items.

Input:
- Conversation, including the candidate assistant response.
- A batch of rubric items with scoring instructions.

Instructions:
- Return exactly one compact machine-readable object and nothing else.
- Do not output markdown fences.
- Do not output hidden reasoning or explanations.
- Return one judgment for every required rubric item.
- Do not use rubric ids such as "r1"; use the provided numeric rubric keys.
- The output is invalid if any required rubric item is missing.

For each rubric item:
- If the item uses a 1--10 scale, assign a numeric score from 1 to 10 according
  to the rubric.
- If the item uses partial points, assign the number of awarded points from 0 to
  the maximum points.
- If the item contains explicit scoring subitems, judge each subitem
  independently and return either its satisfaction probability or its awarded
  points, depending on the scoring mode.

The reward code normalizes the resulting score to the interval \([0,1]\) before
aggregation.
\end{promptbox}

\end{document}